\documentclass[11pt]{article}
\usepackage{fullpage}
\usepackage{amsfonts,epsfig,graphicx}
\usepackage{amsmath,amssymb,amsthm}
\usepackage{graphics}
\usepackage{macros}
\usepackage[round]{natbib}
\usepackage{color}
\usepackage[ruled]{algorithm2e}
\usepackage{epstopdf}
\usepackage{algorithmic}
\usepackage{enumerate}
\usepackage[bookmarks=true,colorlinks,citecolor=blue,urlcolor=blue]{hyperref}

\newtheorem{lemma}{Lemma}
\newtheorem{theorem}{Theorem}
\newtheorem{proposition}{Proposition}
\newtheorem*{definition}{Definition}
\newtheorem{corollary}{Corollary}

\newenvironment{carlist}
 {\begin{list}{$\bullet$}
 {\setlength{\topsep}{0in} \setlength{\partopsep}{0in}
  \setlength{\parsep}{0in} \setlength{\itemsep}{\parskip}
  \setlength{\leftmargin}{0.07in} \setlength{\rightmargin}{0.08in}
  \setlength{\listparindent}{0in} \setlength{\labelwidth}{0.08in}
  \setlength{\labelsep}{0.1in} \setlength{\itemindent}{0in}}}
 {\end{list}}

\newcommand{\bcar}{\begin{carlist}}
\newcommand{\ecar}{\end{carlist}}

\newcommand{\myparagraph}[1]{\subsubsection{#1}}
\newcommand{\lhil}[1]{\ensuremath{\|#1\|_H}}

\newcommand{\elltil}{\ensuremath{\widetilde{\ell}}}
\newcommand{\wtil}{\ensuremath{\widetilde{w}}}

\newenvironment{compactenumerate}
{ \begin{enumerate}
    \setlength{\itemsep}{1pt}
    \setlength{\parskip}{1pt}
    \setlength{\parsep}{1pt}     }
{ \end{enumerate}                  }

\title{Learning Halfspaces and Neural Networks with\\ Random Initialization}
\author{Yuchen Zhang \quad Jason D. Lee \quad Martin J. Wainwright \quad Michael I. Jordan\\\\
Department of Electrical Engineering and Computer Science\\ University
of California, Berkeley, CA 94709\\
\tt \{yuczhang,jasondlee88,wainwrig,jordan\}@eecs.berkeley.edu}
\date{}

\begin{document}

\maketitle

\begin{abstract}
\noindent We study non-convex empirical risk minimization for learning halfspaces and neural networks. For loss functions that are $L$-Lipschitz continuous, we present algorithms to learn halfspaces and multi-layer neural networks that achieve arbitrarily small excess risk $\epsilon>0$. The time complexity is polynomial in the input dimension $d$ and the sample size $n$, but exponential in the quantity $(L/\epsilon^2)\log(L/\epsilon)$. These algorithms run multiple rounds of random initialization followed by arbitrary optimization steps. We further show that if the data is separable by some neural network with constant margin $\gamma>0$, then there is a polynomial-time algorithm for learning a neural network that separates the training data with margin $\Omega(\gamma)$. As a consequence, the algorithm achieves arbitrary generalization error $\epsilon>0$ with $\poly(d,1/\epsilon)$ sample and time complexity. We establish the same learnability result when the labels are randomly flipped with probability $\eta<1/2$.
\end{abstract}

\section{Introduction}

The learning of a halfspace is the core problem solved by many machine
learning methods, including the
Perceptron~\citep{rosenblatt1958perceptron}, the Support Vector
Machine~\citep{vapnik1998statistical} and
AdaBoost~\citep{freund1997decision}. More formally, for a given input
space $\mathcal{X} \subset \R^d$, a halfspace is defined by a linear
mapping $f(x) = \langle w, x \rangle$ from $\mathcal{X}$ to the real
line.  The sign of the function value $f(x)$ determines if $x$ is
located on the positive side or the negative side of the halfspace. A
labeled data point consists of a pair $(x,y) \in \mathcal{X} \times
\{-1, 1\}$, and given $n$ such pairs $\{(x_i,y_i)\}_{i=1}^n$, the
empirical prediction error is given by
\begin{align}
\label{eqn:f-zero-one-loss}
\ell(f) \defeq \frac{1}{n}\sum_{i=1}^n \indicator[- y_if(x_i) \geq 0 ].
\end{align}
The loss function in equation~\eqref{eqn:f-zero-one-loss} is also
called the zero-one loss. In agnostic learning, there is no hyperplane
which perfectly separates the data, in which case the goal is to find
a mapping $f$ that achieves a small zero-one loss.  The method of
choosing a function $f$ based on minimizing the
criterion~\eqref{eqn:f-zero-one-loss} is known as empirical risk
minimization (ERM).

It is known that finding a halfspace that approximately minimizes the
zero-one loss is NP-hard. In particular, \cite{guruswami2009hardness} show that, for any $\epsilon
\in (0, 1/2]$, given a set of pairs such that the optimal zero-one
  loss is bounded by $\epsilon$, it is NP-hard to find a halfspace
  whose zero-one loss is bounded by $1/2 - \epsilon$. Many practical
  machine learning algorithms minimize convex surrogates of the
  zero-one loss, but the halfspaces obtained through the convex
  surrogate are not necessarily optimal. In fact, the result of \cite{guruswami2009hardness} shows that
  the approximation ratio of such procedures could be arbitrarily
  large.

In this paper, we study optimization problems of the form
\begin{align}
\label{eqn:f-general-h-loss}
\ell(f) \defeq \frac{1}{n}\sum_{i=1}^n h(- y_if(x_i)),
\end{align}
where the function $h: \R \rightarrow \R$ is $L$-Lipschitz continuous
for some $L < +\infty$, but is otherwise arbitrary (and so can be
nonconvex). This family does not include the zero-one-loss (since it
is not Lipschitz), but does include functions that can be used to
approximate it to arbitrary accuracy with growing $L$.  For instance,
the piecewise-linear function:
\begin{align}
\label{eqn:approx-zero-one-h}
h(x) \defeq \begin{cases} 0 & x\leq -\frac{1}{2L},\\ 
1 & x\geq \frac{1}{2L},\\ Lx + 1/2 & \mbox{otherwise},
\end{cases}
\end{align}
is $L$-Lipschitz, and converges to the step function as the parameter
$L$ increases to infinity.  

\cite{shalev2011learning} study the problem of
minimizing the objective~\eqref{eqn:f-general-h-loss} with function $h$ defined by~\eqref{eqn:approx-zero-one-h}, and show that
under a certain cryptographic assumption, there is no $\poly(L)$-time
algorithm for (approximate) minimization.  Thus, it is reasonable to
assume that the Lipschitz parameter $L$ is a constant that does not
grow with the dimension $d$ or sample size $n$.  Moreover, when $f$ is
a linear mapping, scaling the input vector $x$ or scaling the weight
vector $w$ is equivalent to scaling the Lipschitz constant. Thus, we
assume without loss of generality that the norms of $x$ and $w$ are
bounded by one.

\subsection{Our contributions}

The first contribution of this paper is to present two
$\poly(n,d)$-time methods---namely,
Algorithm~\ref{alg:linear-model-l2l2} and
Algorithm~\ref{alg:linear-model-linfl1}---for minimizing the cost
function~\eqref{eqn:f-general-h-loss} for an arbitrary $L$-Lipschitz
function $h$. We prove that for any given tolerance $\epsilon > 0$,
these algorithms achieve an $\epsilon$-excess risk by running multiple
rounds of random initialization followed by a constant number of
optimization rounds (e.g.,~using an algorithm such as stochastic
gradient descent). The first algorithm is based on choosing the
initial vector uniformly at random from the Euclidean sphere; despite
the simplicity of this scheme, it still has non-trivial guarantees.
The second algorithm makes use a better initialization obtained by
solving a least-squares problem, thereby leading to a stronger
theoretical guarantee.  Random initialization is a widely used
heuristic in non-convex ERM; our analysis supports this usage but
suggests that a careful theoretical treatment of the initialization
step is necessary.

Our algorithms for learning halfspaces have running time that grows
polynomially in the pair $(n, d)$, as well as a term proportional to
$\exp((L/\epsilon^2)\log(L/\epsilon))$. Our next contribution is to
show that under a standard complexity-theoretic assumption---namely,
that $\bf RP \neq NP$---this exponential dependence on $L/\epsilon$
cannot be avoided.  More precisely, letting $h$ denote the piecewise
linear function from equation~\eqref{eqn:approx-zero-one-h} with $L =
1$, Proposition~\ref{theorem:piecewise-linear-hard} shows that there
is no algorithm achieving arbitrary excess risk $\epsilon > 0$ in
$\poly(n,d,1/\epsilon)$ time when $\bf RP \neq NP$. Thus, the random
initialization scheme is unlikely to be substantially improved.

We then extend our approach to the learning of multi-layer neural
networks, with a detailed analysis of the family of $m$-layer
sigmoid-activated neural networks, under the assumption that
$\ell_1$-norm of the incoming weights of any neuron is assumed to be
bounded by a constant $B$. We specify a method
(Algorithm~\ref{alg:nn-rand}) for training networks over this family,
and in Theorem~\ref{theorem:nn-rand}, we prove that its loss is at
most an additive term of $\epsilon$ worse than that of the best neural
network. The time complexity of the algorithm scales as
$\poly(n,d,C_{m,B,1/\epsilon})$, where the constant
$C_{m,B,1/\epsilon}$ does not depends on the input dimension or the
data size, but may depend exponentially on the triplet
$(m,B,1/\epsilon)$.

Due to the exponential dependence on $1/\epsilon$, this agnostic
learning algorithm is too expensive to achieve a diminishing excess
risk for a general data set.  However, by analyzing data sets that are
separable by some neural network with constant margin $\gamma > 0$, we
obtain a stronger achievability result.  In particular, we show in
Theorem~\ref{theorem:separable-data-error} that there is an efficient
algorithm that correctly classifies all training points with margin
$\Omega(\gamma)$ in polynomial time. As a consequence, the algorithm
learns a neural network with generalization error bounded by
$\epsilon$ using $\poly(d,1/\epsilon)$ training points and in
$\poly(d,1/\epsilon)$ time. This so-called \emph{BoostNet} algorithm
uses the AdaBoost approach~\citep{freund1997decision} to construct a
$m$-layer neural network by taking an $(m-1)$-layer network as a weak
classifier. The shallower networks are trained by the agnostic
learning algorithms that we develop in this paper. We establish the
same learnability result when the labels are randomly flipped with
probability $\eta < 1/2$ (see Corollary~\ref{coro:flip-label}). Although the time complexity of BoostNet is
exponential in $1/\gamma$, we demonstrate that our achievable result
is unimprovable---in particular, by showing that a
$\poly(d,1/\epsilon,1/\gamma)$ complexity is impossible under a
certain cryptographic assumption (see
Proposition~\ref{theorem:hardness-neuralnet-margin}).

Finally, we report experiments on learning parity functions with
noise, which is a challenging problem in computational learning
theory. We train two-layer neural networks using
BoostNet, then compare them with the traditional backpropagation
approach. The experiment shows that BoostNet learns the degree-5
parity function by constructing 50 hidden neurons, while the
backpropagation algorithm fails to outperform random guessing.


\subsection{Related Work}

This section is devoted to discussion of some related work so as to
put our contributions into broader context.

\myparagraph{Learning halfspaces} The problem of learning halfspaces
is an important problem in theoretical computer science. It is known
that for any constant approximation ratio, the problem of
approximately minimizing the zero-one loss is computationally
hard~\citep{guruswami2009hardness,daniely2014average}. Halfspaces can
be efficiently learnable if the data are drawn from certain special
distributions, or if the label is corrupted by particular forms of
noise. Indeed, \cite{blum1998polynomial} and \cite{servedio2001efficient} show that if the labels are
corrupted by random noise, then the halfspace can be learned in
polynomial time. The same conclusion was established by \cite{awasthi2015efficient} when the labels are corrupted by
Massart noise, and the covariates are drawn from the uniform
distribution on a unit sphere. When the label noise is adversarial,
the halfspace can be learned if the data distribution is isotropic
log-concave and the fraction of labels being corrupted is bounded by a
small
quantity~\citep{kalai2008agnostically,klivans2009learning,awasthi2014power}. When
no assumption is made on the noise, \cite{kalai2008agnostically} show that if the data are drawn from
the uniform distribution on a unit sphere, then there is an algorithm
whose time complexity is polynomial in the input dimension, but
exponential in $1/\epsilon$ (where $\epsilon$ is the additive
error). In this same setting, \cite{klivans2014embedding} prove that the exponential
dependence on $1/\epsilon$ is unavoidable.

Another line of work modifies the loss function to make it easier to
minimize. \cite{simon2001efficient} suggest
comparing the zero-one loss of the learned halfspace to the optimal
$\mu$-margin loss. The $\mu$-margin loss asserts that all points whose
classification margins are smaller than $\mu$ should be marked as
misclassified. Under this metric, it was shown by \cite{simon2001efficient,birnbaum2012learning} that the optimal
$\mu$-margin loss can be achieved in polynomial time if $\mu$ is a
positive constant. \cite{shalev2011learning}
study the minimization of a continuous approximation to the zero-one
loss, which is similar to our setup. They propose a kernel-based
algorithm which performs as well as the best linear
classifier. However, it is an improper learning method in that the
classifier cannot be represented by a halfspace.

\myparagraph{Learning neural networks}

It is known that any smooth function can be approximated by a neural
network with just one hidden layer~\citep{barron1993universal}, but
that training such a network is NP-hard~\citep{blum1992training}.  In
practice, people use optimization algorithms such as stochastic
gradient (SG) to train neural networks. Although strong theoretical
results are available for SG in the setting of convex objective
functions, there are few such results in the nonconvex setting of
neural networks.

Several recent papers address the challenge of establishing
polynomial-time learnability results for neural networks. \cite{arora2013provable} study the recovery of denoising
auto-encoders. They assume that the top-layer values of the network
are randomly generated and all network weights are randomly drawn from
$\{-1,1\}$. As a consequence, the bottom layer generates a sequence of
random observations from which the algorithm can recover the network
weights. The algorithm has polynomial-time complexity and is capable
of learning random networks that are drawn from a specific
distribution. However, in practice people want to learn deterministic
networks that encode data-dependent representations.

\cite{sedghi2014provable} study the supervised
learning of neural networks under the assumption that the score
function of the data distribution is known. They show that if the
input dimension is large enough and the network is sparse enough, then
the first network layer can be learned by a polynomial-time
algorithm. More recently, \cite{janzamin2015generalization} propose another algorithm relying on the score function that
removes the restrictions of \cite{sedghi2014provable}. The assumption in this case is
that the network weights satisfy a non-degeneracy condition; however,
the algorithm is only capable of learning neural networks with one
hidden layer. Our algorithm does not impose any assumption on the data
distribution, and is able to learn multi-layer neural networks.

Another approach to the problem is via the improper learning
framework. The goal in this case is to find a predictor that is not a
neural network, but performs as well as the best possible neural
network in terms of the generalization error. \cite{livni2014computational} propose a polynomial-time algorithm
to learn networks whose activation function is quadratic. \cite{zhang2015ell_1} propose an algorithm for improper learning
of sigmoidal neural networks. The algorithm runs in
$\poly(n,d,1/\epsilon)$ time if the depth of the networks is a
constant and the $\ell_1$-norm of the incoming weights of any node is
bounded by a constant. It outputs a kernel-based classifier while our
algorithm outputs a proper neural network. On the other hand, in the
agnostic setting, the time complexity of our algorithm depends
exponentially on $1/\epsilon$.


\section{Preliminaries} 

In this section, we formalize the problem set-up and present several
preliminary lemmas that are useful for the theoretical analysis.
We first set up some notation so as to define a general empirical risk
minimization problem. Let $\mathcal{D}$ be a dataset containing $n$
points $\{(x_i,y_i)\}_{i=1}^n$ where $x_i\in \mathcal{X}\subset
\R^d$ and $y_i\in \{-1,1\}$. To goal is to learn a function $f:
\mathcal{X}\to \R$ so that $f(x_i)$ is as close to $y_i$ as
possible. We may write the loss function as
\begin{align}
\label{eqn:loss-ell-f}
\ell(f) \defeq \sum_{i=1}^n \alpha_i h(-y_i f(x_i)).
\end{align}
\begin{figure}
\begin{tabular}{ccc}
\includegraphics[width = 0.3\textwidth]{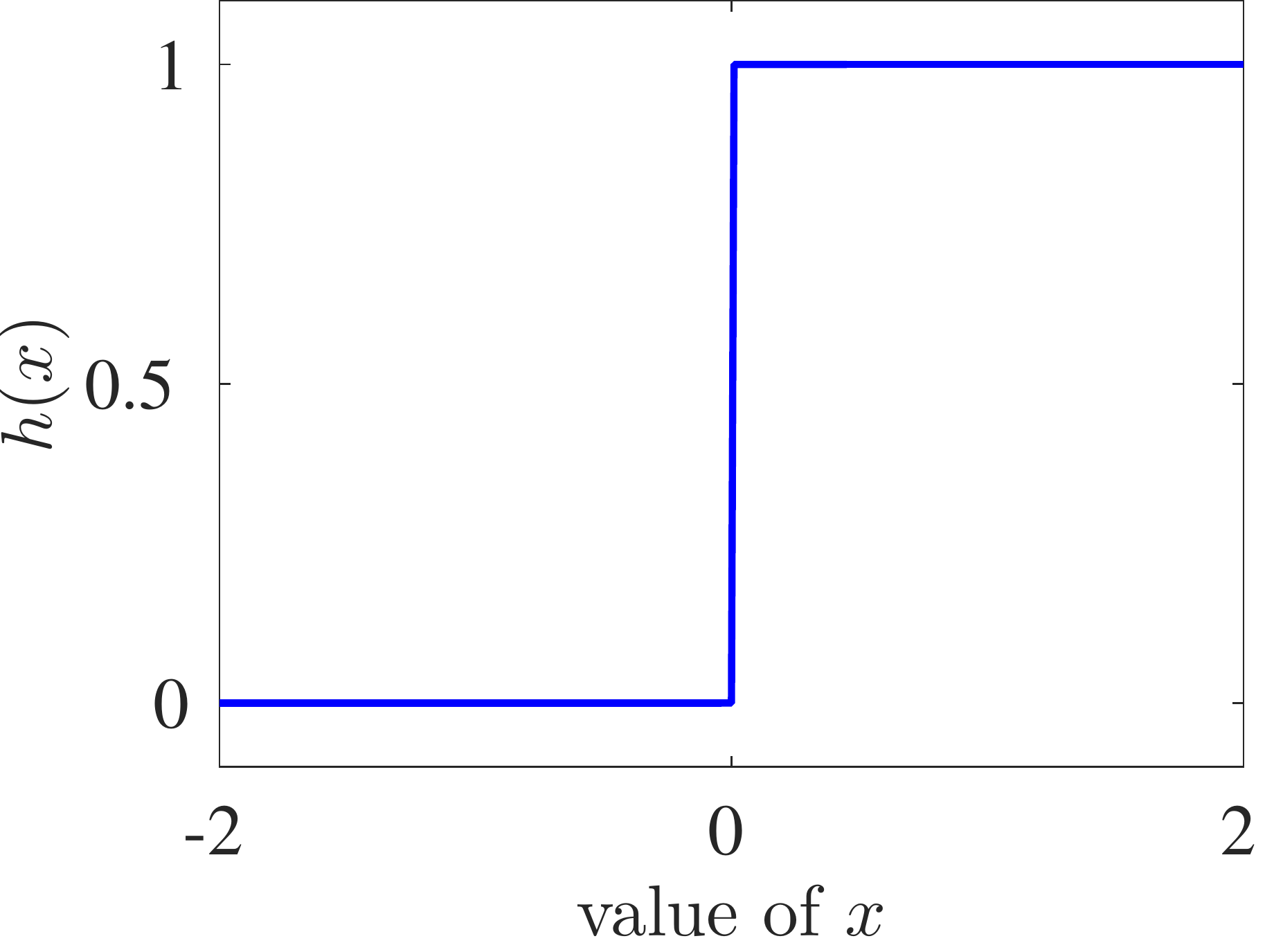} &
\includegraphics[width = 0.3\textwidth]{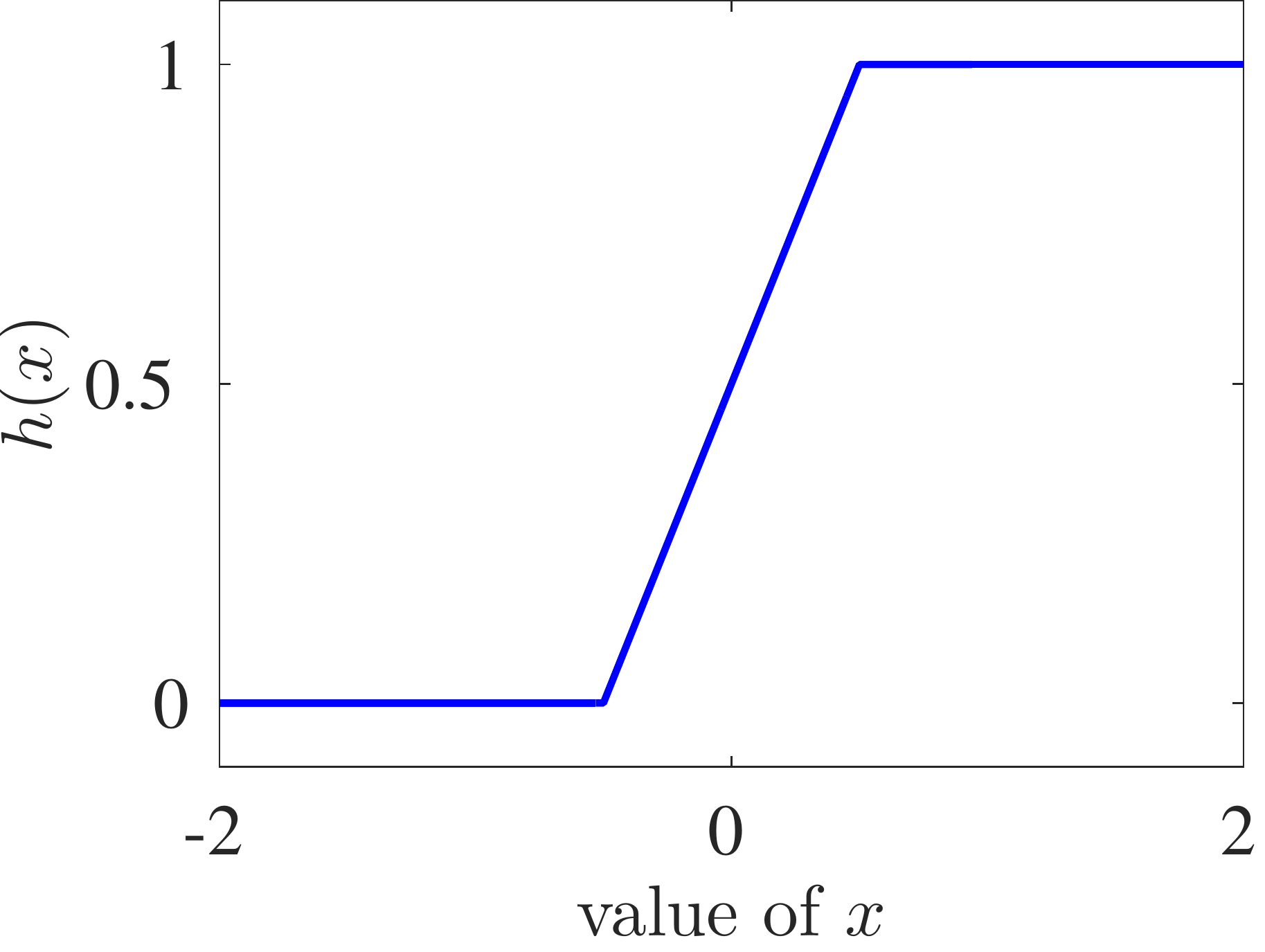} &
\includegraphics[width = 0.3\textwidth]{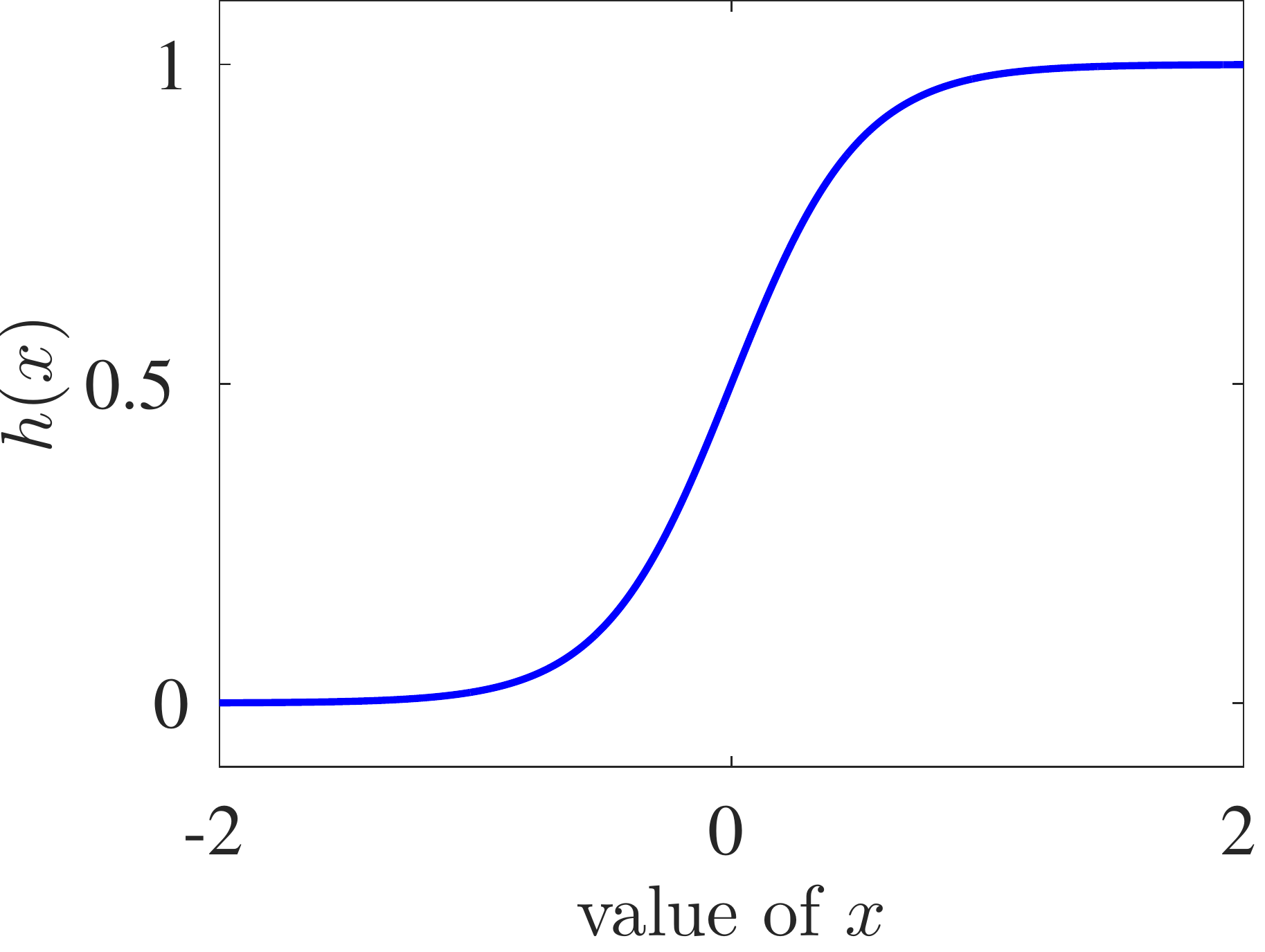}\\
(a) Step function & (b) Piecewise linear function & (c) Sigmoid
function
\end{tabular}
\caption{Comparing the step function (panel (a)) with its two
  continuous approximations (panels (b) and (c)).}
\label{fig:h-func}
\end{figure}
where $h_i:\R\to\R$ is a $L$-Lipschitz continuous function that
depends on $y_i$, and $\{\alpha_1,\dots,\alpha_n\}$ are non-negative
importance weights that sum to $1$. As concrete examples, the function
$h$ can be the piecewise linear function defined by
equation~\eqref{eqn:approx-zero-one-h} or the sigmoid function $h(x)
\defeq 1/(1+e^{- 4 L x})$. 
Figure~\ref{fig:h-func} compares the step function $\indicator[x\geq
0]$ with these continuous approximations.  In the following sections,
we study minimizing the loss function in
equation~\eqref{eqn:loss-ell-f} when $f$ is either a linear mapping or
a multi-layer neural network.

Let us introduce some useful shorthand notation.  We use $[n]$ to
denote the set of indices $\{1,2,\dots,n\}$.  For $q \in [1, \infty]$,
let $\norms{x}_q$ denote the $\ell_q$-norm of vector $x$, given by
$\norms{x}_q \defeq (\sum_{j=1}^d x_j^q)^{1/q}$, as well as
$\norms{q}_\infty = \max \limits_{j = 1, \ldots, d} |x_j|$.  If $u$ is
a $d$-dimensional vector and $\sigma:\R\to\R$ is a function, we use
$\sigma(u)$ as a convenient shorthand for the vector
$(\sigma(u_1),\dots,\sigma(u_d))$.  Given a class $\mathcal{F}$ of
real-valued functions, we define the new function class $\sigma\circ
\mathcal{F} \defeq \big \{ \sigma \circ f \mid f\in \mathcal{F} \big
\}$.


\paragraph{Rademacher complexity bounds:} 

Let $\{(x'_j,y'_j)\}_{j=1}^k$ be i.i.d.~samples drawn from the dataset
$\mathcal{D}$ such that probability of drawing $(x_i,y_i)$ is
proportional to $\alpha_i$. We define the sample-based loss function:
\begin{align}
\label{eqn:sample-based-loss}
G(f) \defeq \frac{1}{k} \sum_{j=1}^k h(-y'_jf(x'_j)).
\end{align} 
It is straightforward to verify that $\E[G(f)] = \ell(f)$.  For a
given function class $\mathcal{F}$, the Rademacher complexity of
$\mathcal{F}$ with respect to these $k$ samples is defined as
\begin{align}
\label{eqn:define-rademacher-complexity}
R_k(\mathcal{F}) \defeq \E\Big[ \sup_{f\in \mathcal{F}} \frac{1}{k}
  \sum_{j=1}^k \varepsilon_j f(x'_j) \Big],
\end{align}
where the $\{\varepsilon_j\}$ are independent Rademacher random
variables.

\begin{lemma}\label{lemma:rademacher-generalization}
Assume that $\mathcal{F}$ contains the constant zero function $f(x) \equiv 0$, then we have 
\[
\E\Big[ \sup \limits_{f\in \mathcal{F}} |G(f)-\ell(f)|\Big] \leq 4L
R_k(\mathcal{F}).\]
\end{lemma}

\noindent This lemma shows that $R_k(\mathcal{F})$ controls the
distance between $G(f)$ and $\ell(f)$. For the function classes
studied in this paper, we will have $R_k(\mathcal{F}) =
\order(1/\sqrt{k})$. Thus, the function $G(f)$ will be a good
approximation to $\ell(f)$ if the sample size is large enough. This
lemma is based on a slight sharpening of the usual Ledoux-Talagrand
contraction for Rademacher variables~\citep{Ledoux2013probability}; see
Appendix~\ref{sec:proof-rademacher-generalization} for the proof.

\paragraph{Johnson-Lindenstrauss lemma:} 

The Johnson-Lindenstrauss lemma is a useful tool for dimension
reduction. It gives a lower bound on an integer $s$ such that after
projecting $n$ vectors from a high-dimensional space into an
$s$-dimensional space, the pairwise distances between this collection
of vectors will be approximately preserved.

\begin{lemma}
\label{lemma:jl}
For any $0 < \epsilon < 1/2$ and any positive integer $n$, consider
any positive integer \mbox{$s \geq 12 \log n/\epsilon^2$.} Let $\phi$
be a operator that projects a vector in $\R^d$ to a random
$s$-dimensional subspace of $\R^d$, then scales the vector by
$\sqrt{d/s}$. Then for any set of vectors $u_1,\dots,u_n\in \R^d$, we
have
\begin{align}
\Big| \ltwos{u_i - u_j}^2 - \ltwos{\phi(u_i) - \phi(u_j)}^2 \Big| \leq
\epsilon \ltwos{u_i - u_j}^2 \quad \mbox{for every $i,j\in[n]$}.
\end{align}
holds with probability at least $1/n$.
\end{lemma}
\noindent See the paper by \cite{dasgupta1999elementary} for a simple
proof.

\paragraph{Maurey-Barron-Jones lemma:} 
Letting $G$ be a subset of any Hilbert space $H$, the
Maurey-Barron-Jones lemma guarantees that any point $v$ in the convex
hull of $G$ can be approximated by a convex combination of a small
number of points of $G$.  More precisely, we have:

\begin{lemma}
\label{lemma:mbj}
Consider any subset $G$ of any Hilbert space such that $\lhil{g} \leq
b$ for all $g\in G$. Then for any point $v$ is in the convex hull of
$G$, there is a point $v_s$ in the convex hull of $s$ points of $G$
such that $\lhil{v - v_s}^2 \leq b^2/s$.
\end{lemma}
\noindent See the paper by \cite{pisier1980remarques} for a
proof. This lemma is useful in our analysis of neural network
learning.


\section{Learning Halfspaces}

In this section, we assume that the function $f$ is a linear mapping
$f(x)\defeq \langle w, x\rangle$, so that our cost function can be
written as
\begin{align}
\label{eqn:define-linear-loss}
\ell(w) \defeq \sum_{i=1}^n \alpha_i h(-y_i\langle w, x_i\rangle).
\end{align}
In this section, we present two polynomial-time algorithms to
approximately minimize this cost function over certain types of
$\ell_p$-balls.  Both algorithms run multiple rounds of random
initialization followed by arbitrary optimization steps. The first
algorithm initializes the weight vector by uniformly drawing from a
sphere.  We next present and analyze a second algorithm in which the
initialization follows from the solution of a least-square problem. It
attains the stronger theoretical guarantees promised in the
introduction.


\subsection{Initialization by uniform sampling}
\label{sec:halfspace-algorithm-1} 

We first analyze a very simple algorithm based on uniform
sampling. Letting $\wstar$ denote a minimizer of the objective
function~\eqref{eqn:define-linear-loss}, by rescaling as necessary, we
may assume that $\ltwos{\wstar} = 1$.  As noted earlier, by redefining
the function $h$ as necessary, we may also assume that the points
$\{x_i\}_{i=1}^n$ all lie inside the Eulcidean ball of radius
one---that is, $\ltwos{x_i} \leq 1$ for all $i \in [\numobs]$.

Given this set-up, a simply way in which to estimate $\wstar$ is to
first draw $w$ uniformly from the Euclidean unit sphere, and then
apply an iterative scheme to minimize the the loss function with the
randomly drawn vector $w$ as the initial point.  At an intuitive
level, this algorithm will find the global optimum if the initial
weight is drawn sufficiently close to $\wstar$, so that the iterative
optimization method converges to the global minimum. However, by
calculating volumes of spheres in high-dimensions, it could require
$\Omega(\frac{1}{\epsilon^d})$ rounds of random sampling before
drawing a vector that is sufficiently close to $\wstar$, which makes
it computationally intractable unless the dimension $d$ is small.

In order to remove the exponential dependence,
Algorithm~\ref{alg:linear-model-l2l2} is based on drawing the initial
vector from a sphere of radius $r > 1$, where the radius $r$ should be
viewed as hyper-parameter of the algorithm. One should choose a
greater $r$ for a faster algorithm, but with a less accurate solution.
The following theorem characterizes the trade-off between the accuracy
and the time complexity. 

\begin{algorithm}[t] \DontPrintSemicolon \KwIn{Feature-label
pairs $\{(x_i,y_i)\}_{i=1}^n$; feature dimension $d$; parameters $T$
and $r \geq 1$.}  \vspace{5pt} For $t =
1,2,\dots,T$: \begin{enumerate} \item Draw a random vector $u$
uniformly from the unit sphere of $\R^d$.  \item Choose $w_t \defeq
ru$, or compute $w_t$ by a $\poly(n,d)$-time algorithm such that
$\ltwos{w_t} \leq r$ and $\ell(w_t) \leq \ell(ru)$.  \end{enumerate}
\KwOut{$\what \defeq \arg\min_{w\in \{w_1,\dots,w_T\}}
\ell(w)$.}  
\caption{Learning halfspaces based on initializing from uniform sampling} \label{alg:linear-model-l2l2} 
\end{algorithm}

\begin{theorem}\label{theorem:bound-l2l2}
For $0 < \epsilon <1/2$ and $\delta > 0$, let $s \defeq
\min\{d,\lceil 12\log(n+2)/\epsilon^2 \rceil\}$, $r\defeq \sqrt{d/s}$
and $T \defeq \lceil (2n+4)(\pi/\epsilon)^{s-1}\log(1/\delta)\rceil$.
With probability at least $1-\delta$,
Algorithm~\ref{alg:linear-model-l2l2} outputs a vector $\ltwos{\what}
\leq r$ which satisfies: \[ \ell(\what) \leq \ell(\wstar) + 6
\epsilon L.  \] The time complexity is bounded by
$\poly(n^{(1/\epsilon^2)\log^2(1/\epsilon)},d,
\log(1/\delta))$. 
\end{theorem}  

The proof of Theorem~\ref{theorem:bound-l2l2}, provided in
Appendix~\ref{sec:proof-theorem-l2l2}, uses the Johnson-Lindenstrauss
lemma. More specifically, suppose that we project the weight vector
$w$ and all data points $x_1,\dots,x_n$ to a random subspace $S$ and
properly scale the projected vectors.  The Johnson-Lindenstrauss lemma
then implies that with a constant probability, the inner products
$\langle w, x_i
\rangle$ will be almost invariant after the projection---that is,
\begin{align*}
\langle \wstar, x_i \rangle \approx \langle \phi(\wstar), \phi(x_i)
\rangle = \langle r\phi(\wstar), x_i \rangle \quad \mbox{for
  every $i\in[n]$},
\end{align*}
where $r$ is the scale factor of the projection. As a consequence, the
vector $r\phi(\wstar)$ will approximately minimize the loss.  If we
draw a vector $v$ uniformly from the sphere of $S$ with radius $r$ and
find that $v$ is sufficiently close to $r\phi(\wstar)$, then we call
this vector $v$ as a successful draw. The probability of a successful
draw depends on the dimension of $S$, independent of the original
dimension~$d$ (see Lemma~\ref{lemma:jl}). If the draw is successful,
then we use $v$ as the initialization so that it approximately
minimize the loss. Note that drawing from the unit sphere of a random
subspace $S$ is equivalent to directly drawing from the original space
$\R^d$, so that there is no algorithmic need to explicitly construct
the random subspace.

It is worthwhile to note some important deficiencies of
Algorithm~\ref{alg:linear-model-l2l2}. First, the algorithm outputs a
vector satisfying the bound $\ltwos{\what}\leq r$ with $r > 1$, so that
it is not guaranteed to lie within the Euclidean unit ball. Second,
the $\ell_2$-norm constraints $\ltwos{x_i}\leq 1$ and $\ltwos{\wstar}
= 1$ cannot be generalized to other norms. Third, the complexity term
$n^{(1/\epsilon^2)\log^2(1/\epsilon)}$ has a power growing
with~$1/\epsilon$.  Our second algorithm overcomes these limitations.


\subsection{Initialization by solving a least-square problem}

\begin{algorithm}[t]
\DontPrintSemicolon \KwIn{Feature-label pairs $\{(x_i,y_i)\}_{i=1}^n$;
  feature dimension $d$; parameters $k,T$.}
\vspace{5pt}
For $t = 1,2,\dots,T$:
\begin{compactenumerate}
\item Sample $k$ points $\{(x'_j,y'_j)\}_{j=1}^k$ from the dataset
  with respect to their importance weights. Sample a vector $u$
  uniformly from $[-1,1]^k$ and let $v \defeq \argmin_{\norms{w}_p\leq
    1} \sum_{j=1}^k (\langle w, x'_j \rangle - u_j)^2$.
\item Choose $w_t \defeq v$, or compute $w_t$ by a $\poly(n,d)$-time
  algorithm such that $\norms{w_t}_p \leq 1$ and $\ell(w_t) \leq
  \ell(v)$.
\end{compactenumerate}
\KwOut{$\what \defeq \arg\min_{w\in \{w_1,\dots,w_T\}} \ell(w)$.}
\caption{Learning halfspaces based on initializing from the solution
  of a least-squares problem.}
\label{alg:linear-model-linfl1}
\end{algorithm}

We turn to a more general setting in which $\wstar$ and $x_i$ are
bounded by a general $\ell_q$-norm for some $q \in [2, \infty]$.
Letting $p \in [1,2]$ denote the associated dual exponent ( i.e., such
that $1/p + 1/q = 1$), we assume that $\norms{\wstar}_p \leq 1$ and
$\norms{x_i}_q \leq 1$ for every $i\in[n]$.  Note that this
generalizes our previous set-up, which applied to the case $p = q =
2$.  In this setting, Algorithm~\ref{alg:linear-model-linfl1} is a
procedure that outputs an approximate minimizer of the loss function.
In each iteration, it draws $k \ll n$ points from the data set
$\mathcal{D}$, and then constructs a random least-squares problem
based on these samples. The solution to this problem is used to
initialize an optimization step.

The success of Algorithm~\ref{alg:linear-model-linfl1} relies on the
following observation: if we sample $k$ points independently from the
dataset, then Lemma~\ref{lemma:rademacher-generalization} implies that
the sample-based loss $G(f)$ will be sufficiently close to the
original loss $\ell(f)$. Thus, it suffices to minimize the
sample-based loss. Note that $G(f)$ is uniquely determined by the $k$
inner products $\varphi(w) \defeq (\langle w, x_{i_1} \rangle, \dots,
\langle w, x_{i_k} \rangle)$. If there is a vector $w$ satisfying
$\varphi(w) = \varphi(\wstar)$, then its performance on the
sample-based loss will be equivalent to that of $\wstar$. As a
consequence, if we draw a vector $u\in [-1,1]^k$ that is sufficiently
close to $\varphi(\wstar)$ (called a successful $u$), then we can
approximately minimize the sample-based loss by solving $\varphi(w) =
u$, or alternatively by minimizing $\ltwos{\varphi(w)-u}^2$. The
latter problem can be solved by a convex program in polynomial
time. The probability of drawing a successful $u$ only depends on $k$,
independent of the input dimension and the sample size. This allows
the time complexity to be polynomial in~$(n,d)$. The trade-off between
the target excess risk and the time complexity is characterized by the
following theorem.  For given $\epsilon,\delta > 0$, it is based on
running Algorithm~\ref{alg:linear-model-linfl1} with the choices
\begin{align}
\label{eqn:convex-alg-choose-k}
k \defeq \begin{cases} \lceil 2 \log d / \epsilon^2 \rceil & \mbox{if
  }p = 1\\ 
\lceil (q-1)/\epsilon^2 \rceil & \mbox{if }p > 1,
\end{cases}
\end{align}
and $T \defeq \lceil 5(4/\epsilon)^k\log(1/\delta) \rceil$.

\begin{theorem}\label{theorem:bound-linfl1}
For given $\epsilon, \delta > 0$, with the choices of $(k, T)$ given
above, Algorithm~\ref{alg:linear-model-linfl1} outputs a vector
$\norms{\what}_p \leq 1$ such that
\begin{align*}
\ell(\what) \leq \ell(\wstar) + 11 \epsilon L \qquad \mbox{with
  probability at least $1-\delta$.}
\end{align*}
The time complexity is bounded by 
\begin{align*}
\begin{cases}
        \poly \big (n,d^{(1/\epsilon^2)\log(1/\epsilon)}, \log(1/\delta) \big)
        & \mbox{if } p = 1\\
\poly \big(n,d,e^{(q/\epsilon^2)\log(1/\epsilon)}, \log(1/\delta) \big) &
\mbox{if } p > 1.
\end{cases}
\end{align*}
\end{theorem}

\noindent 
See Appendix~\ref{sec:proof-bound-linfl1} for the proof.
Theorem~\ref{alg:linear-model-linfl1} shows that the time complexity
of the algorithm has polynomial dependence on $(n,d)$ but exponential
dependence on $L/\epsilon$. \cite{shalev2011learning} proved a similar complexity bound when
the function $h$ takes the piecewise-linear
form~\eqref{eqn:approx-zero-one-h}, but our algorithm applies to
arbitrary continuous functions. We note that the result is interesting
only when $(L/\epsilon)^2 \ll d$, since otherwise the same time
complexity can be achieved by a grid search of $\wstar$ within the
$d$-dimensional unit ball.


\subsection{Hardness result}

In Theorem~\ref{alg:linear-model-linfl1}, the time complexity has an
exponential dependence on $L/\epsilon$. \cite{shalev2011learning} show that the time complexity cannot be
polynomial in $L$ even for improper learning. It is natural to wonder
if Algorithm~\ref{alg:linear-model-linfl1} can be improved to have polynomial dependence on
$(n,d,1/\epsilon)$ given that $L = 1$. In this section, we provide evidence
that this is unlikely to be the case.

To prove the hardness result, we reduce from the MAX-2-SAT problem,
which is known to be NP-hard. In particular, we show that if there is
an algorithm solving the minimization
problem~\eqref{eqn:define-linear-loss}, then it also solves the
MAX-2-SAT problem.  Let us recall the MAX-2-SAT problem:
\begin{definition}[MAX-2-SAT]
Given $n$ literals $\{z_1,\dots,z_n\}$ and $d$ clauses
$\{c_1,\dots,c_d\}$. Each clause is the conjunction of two arguments
that may either be a literal or the negation of a literal \footnote{In
  the standard MAX-2-SAT setup, each clause is the disjunction of two
  literals. However, any disjunction clause can be reduced to three
  conjunction clauses. In particular, a clause $z_1 \vee z_2$ is
  satisfied if and only if one of the following is satisfied: $z_1
  \wedge z_2$, $\neg z_1 \wedge z_2$, $z_1 \wedge \neg z_2$.}. The
goal is to determine the maximum number of clauses that can be
simultaneously satisfied by an assignment.
\end{definition}

Since our interest is to prove a lower bound, it suffices to study a
special case of the general minimization problem---namely, one in
which $\ltwos{\wstar}\leq 1$ and $\ltwos{x_i}\leq 1$, $y_i = - 1$ for
any $i\in[n]$.  The following proposition shows that if $h$ is the
piecewise-linear function with $L = 1$, then approximately minimizing
the loss function is hard. See
Appendix~\ref{sec:proof-piecewise-linear-hard} for the proof. 

\begin{proposition}
\label{theorem:piecewise-linear-hard}
Let $h$ be the piecewise-linear function~\eqref{eqn:approx-zero-one-h}
with Lipschitz constant $L = 1$.  Unless ${\bf RP = NP}$\footnote{$\bf
  RP$ is the class of randomized polynomial-time algorithms.}, there
is no randomized $\poly(n,d,1/\epsilon)$-time algorithm computing a
vector $\what$ which satisfies $\ell(\what) \leq \ell(\wstar) +
\epsilon$ with probability at least $1/2$.
\end{proposition}

\noindent Proposition~\ref{theorem:piecewise-linear-hard} provides a
strong evidence that learning halfspaces with respect to a continuous
sigmoidal loss cannot be done in $\poly(n,d,1/\epsilon)$ time. We note
that \cite{hush1999training} proved a similar hardness result,
but without the unit-norm constraint on $\wstar$ and
$\{x_i\}_{i=1}^n$. The non-convex ERM problem without a unit norm
constraint is notably harder than ours, so this particular hardness
result does not apply to our problem setup.

\section{Learning Neural Networks}

Let us now turn to the case in which the function $f$ represents a
neural network. Given two numbers $p\in (1,2]$ and $q\in [2,\infty)$
such that $1/p + 1/q = 1$, we assume that the input vector satisfies
$\norms{x_i}_q \leq 1$ for every $i\in[n]$. The class of $m$-layer
neural networks is recursively defined in the following way. A
one-layer neural network is a linear mapping from $\R^d$ to $\R$, and
we consider the set of mappings:
\begin{align*}
\nn_1 \defeq \{x\to\langle w, x\rangle:~\norms{w}_p \leq B\}.
\end{align*}
For $m > 1$, an $m$-layer neural network is a linear combination of
$(m-1)$-layer neural networks activated by a sigmoid function, and so
we define:
\begin{align*}
\nn_m \defeq \Big\{x\to \sum_{j=1}^d w_j \sigma(f_j(x)):~ d <
\infty,~f_j\in \nn_{m-1},~\lones{w} \leq B \Big\}.
\end{align*}
In this definition, the function $\sigma: \R\to[-1,1]$ is an arbitrary
1-Lipschitz continuous function. At each hidden layer, we allow the
number of neurons $d$ to be arbitrarily large, but the per-unit $\ell_1$-norm
 must be bounded by a constant $B$. This regularization scheme has been studied by \cite{bartlett1998sample,koltchinskii2002empirical,
bartlett2003rademacher,neyshabur2015norm}.

Assuming a constant $\ell_1$-norm bound might be restrictive for some
applications, but without this norm constraint, the neural network
class activated by any sigmoid-like or ReLU-like function is not
efficiently learnable~\cite[][Theorem 3]{zhang2015ell_1}. On the other hand, the
$\ell_1$-regularization imposes sparsity on the neural network. It is
observed in practice that sparse neural networks are capable of
learning meaningful representations such as by convolutional neural
networks, for instance. Moreover, it has been argued that sparse
connectivity is a natural constraint that can lead to improved
performance in practice~\cite[see, e.g.][]{thom2013sparse}.


\begin{algorithm}[t]
\DontPrintSemicolon \KwIn{Feature-label pairs $\{(x_i,y_i)\}_{i=1}^n$;
  number of layers $m$; parameters $k,s,T,B$.}
\vspace{5pt} For $t = 1,2,\dots,T$:
\begin{compactenumerate}
\item Sample $k$ points $\{(x'_j,y'_j)\}_{j=1}^k$ from the dataset
  with respect to their importance weights.
\item Generate a neural network $g\in \nn_m$ in the following
  recursive way:
\begin{itemize}
\item If $m = 1$, then draw a vector $u$ uniformly from
  $[-B,B]^k$. Let $v \defeq \arg\min_{w\in\R^d: \norms{w}_p\leq B}
  \sum_{j=1}^k (\langle w, x'_j \rangle - u_j)^2$ and return $g: x \to
  \langle v, x \rangle$.
\item If $m > 1$, then generate the $(m-1)$-layer networks
  $g_1,\dots,g_s\in \nn_{m-1}$ using this recursive program. Draw a
  vector $u$ uniformly from $[-B,B]^k$. Let
\vspace{-8pt}
\begin{align}
\label{eqn:nn-actual-problem}
v \defeq \arg\min_{w\in\R^s:\norms{w}_1 \leq
        B} \sum_{j=1}^k \Big(\sum_{l=1}^s w_l \sigma(g_l(x'_j)) -
        u_j\Big)^2
\end{align}
and return $g: x \to \sum_{l=1}^s v_l \sigma(g_l(x))$.
\end{itemize}
\item Choose $f_t \defeq g$, or compute $f_t$ by a $\poly(n,d)$-time
  algorithm such that $f_t\in \nn_m$ and $\ell(f_t) \leq \ell(g)$.
\end{compactenumerate}
\KwOut{$\fhat \defeq \arg\min_{f\in \{f_1,\dots,f_T\}} \ell(f)$.}
\caption{Algorithm for learning neural networks}
\label{alg:nn-rand}
\end{algorithm}


\subsection{Agnostic learning}

In the agnostic setting, it is not assumed there exists a neural
network that separates the data.  Instead, our goal is to compute a
neural network $\fhat\in \nn_m$ that minimizes the loss function over
the space of all given networks.  Letting $\fstar \in \nn_m$ be the
network that minimizes the empirical loss $\ell$, we now present and
analyze a method (see Algorithm~\ref{alg:nn-rand}) that computes a
network whose loss is at most $\epsilon$ worse that that of $\fstar$.
We first state our main guarantee for this algorithm, before providing
intuition.  More precisely, for any $\epsilon,\delta > 0$, the
following theorem applies to Algorithm~\ref{alg:nn-rand} with the
choices:
\begin{align}\label{eqn:setting-k-s-T}
k \defeq
\lceil q/\epsilon^2 \rceil,\quad s \defeq \lceil 1/\epsilon^2 \rceil, \quad T \defeq \left\lceil 5(4/\epsilon)^{k(s^m-1)/(s-1)}
\log(1/\delta)\right\rceil.
\end{align}
\begin{theorem}
\label{theorem:nn-rand}
For given $B \geq 1$ and $\epsilon, \delta > 0$, with the choices of
$(k,s, T)$ given above, Algorithm~\ref{alg:nn-rand} outputs a
predictor $\fhat\in \nn_m$ such that
\begin{align}
\ell(\fhat) \leq \ell(\fstar) + (2m+9) \epsilon L B^m \qquad \mbox{
  with probability at least $1-\delta$.}
\end{align}
The computational complexity is bounded by
$\poly(n,d,e^{q(1/\epsilon^2)^m \log(1/\epsilon)},\log(1/\delta))$.
\end{theorem}

\noindent We remark that if $m=1$, then $\nn_m$ is a class of linear
mappings. Thus, Algorithm~\ref{alg:linear-model-linfl1} can be viewed
as a special case of Algorithm~\ref{alg:nn-rand} for learning
one-layer neural networks.  See Appendix~\ref{sec:proof-nn-rand} for
the proof of Theorem~\ref{theorem:nn-rand}.

The intuition underlying Algorithm~\ref{alg:nn-rand} is similar to
that of Algorithm~\ref{alg:linear-model-linfl1}. Each iteration
involves resampling $k$ independent points from the dataset. By the
Rademacher generalization bound, minimizing the sample-based loss
$G(f)$ will approximately minimize the original loss $\ell(f)$. The
value of $G(f)$ is uniquely determined by the vector $\varphi(f)\defeq
(f(x'_1),\dots,f(x'_k))$. As a consequence, if we draw $u\in [-B,B]^k$
sufficiently close to $\varphi(\fstar)$, then a nearly-optimal neural
network will be obtained by approximately solving~$\varphi(g) \approx
u$, or equivalently $\varphi(g) \approx \varphi(\fstar)$.

In general, directly solving the equation $\varphi(g) \approx u$ would
be difficult even if the vector $u$ were known. In particular, since
our class $\nn_m$ is highly non-linear, solving this equation cannot
be reduced to solving a convex program.  On the other hand, suppose
that we write $\fstar(x) =
\sum_{l=1}^s w_l \sigma(\fstar_l(x))$ for some functions $\fstar_l
\in \nn_m$.  Then the problem becomes much easier if the quantities
$\sigma(\fstar_l(x'_j))$ are already known for every $(j,l)\in
      [k]\times [d]$.  With this perspective in mind, we can
      approximately solve the equation by minimizing
\begin{align}
\label{eqn:nn-ideal-equation}
\min_{w\in \R^d: \lones{w}\leq B}~~ \sum_{j=1}^k \Big(\sum_{l=1}^s w_l
\sigma(\fstar_l(x'_j)) - u_j \Big)^2.
\end{align}
Accordingly, suppose that we draw vectors $a_1,\dots,a_s\in [-B,B]^k$
such that each $a_j$ is sufficiently close to
$\varphi(\fstar_j)$---any such draw is called successful.  We may then
recursively compute $(m-1)$-layer networks $g_1,\dots,g_s$ by first
solving the approximate equation $\varphi(g_l) \approx a_l$, and then
rewriting problem~\eqref{eqn:nn-ideal-equation} as
\begin{align*}
\min_{w\in \R^d: \lones{w}\leq B}~~ \sum_{j=1}^k \Big(\sum_{l=1}^s w_l
\sigma(g_l(x'_j)) - u_j\Big)^2.
\end{align*}
This convex program matches the problem~\eqref{eqn:nn-actual-problem}
in Algorithm~\ref{alg:nn-rand}. Note that the probability of a
successful draw $\{a_1, \ldots, a_s\}$ depends on the dimension
$s$. Although there is no constraint on the dimension of $\nn_m$, the
Maurey-Barron-Jones lemma (Lemma~\ref{lemma:mbj}) asserts that it
suffices to choose $s = \order(1/\epsilon^2)$ to compute an
$\epsilon$-accurate approximation.  We refer the reader to
Appendix~\ref{sec:proof-nn-rand} for the detailed proof.


\begin{algorithm}[t]
\DontPrintSemicolon \KwIn{Feature-label pairs $\{(x_i,y_i)\}_{i=1}^n$;
  number of layers $m \geq 2$; parameters $\delta, \gamma,T,B$.}
\vspace{5pt} Initialize $f_0 = 0$ and $b_0 = 0$. For $t =
1,2,\dots,T$:
\begin{compactenumerate}
\item Define $G_t(g) \defeq \sum_{i=1}^n \alpha_{t,i} \sigma(-y_i
  g(x_i))$ where $g\in \nn_{m-1}$ and $\alpha_{t,i} \defeq
  \frac{e^{-y_i\sigma(f_{t-1} (x_i))}}{\sum_{j=1}^n
    e^{-y_i\sigma(f_{t-1}(x_i))}}$.
\item Compute $\ghat_t \in \nn_{m-1}$ by Algorithm~\ref{alg:nn-rand} such that
\begin{align}\label{eqn:alg-nn-fw-weak-learner}
	G_t(\ghat_t) \leq \inf_{g\in\nn_{m-1}} G_t(g) + \frac{\gamma}{2B}
\end{align}
with probability at least $1 - \delta/T$\footnotemark. Let $\mu_t \defeq
\max\{ -\frac{1}{2}, G_t(\ghat_t)\}$.

\item Set $f_{t} = f_{t-1} + \frac{1}{2}\log(\frac{1 - \mu_t}{1 +
  \mu_t})\ghat_t$ and $b_t = b_{t-1} + \frac{1}{2}\left|\log(\frac{1 -
  \mu_t}{1 + \mu_t})\right|$.
\end{compactenumerate}
\KwOut{$\fhat \defeq \frac{B}{b_T} f_T$.}
\caption{The BoostNet algorithm}
\label{alg:nn-fw}
\end{algorithm}

\subsection{Learning with separable data}

We turn to the case in which the data are separable with a positive
margin.  Throughout this section, we assume that the activation
function of $\nn_m$ is an odd function (i.e., $\sigma(-x) =
-\sigma(x)$).  We say that a given data set $\{(x_i,
y_i)\}_{i=1}^\numobs$ is \emph{separable with margin $\gamma$}, or
$\gamma$-separable for short, if there is a network $\fstar\in \nn_m$
such that $y_i \fstar(x_i) \geq \gamma$ for each $i \in [\numobs]$.
Given a distribution $\mprob$ over the space $\mathcal{X} \times \{-1,
1\}$, we say that it is $\gamma$-separable if there is a network
$\fstar \in \nn_m$ such that $y \fstar(x) \geq \gamma$  almost surely
(with respect to $\mprob$).

Algorithm~\ref{alg:nn-fw} learns a neural network on the separable
data. It uses the AdaBoost approach \citep{freund1997decision} to construct the
network, and we refer to it as the \emph{BoostNet algorithm}. In each
iteration, it trains a weak classifier $\ghat_t \in \nn_{m-1}$ with an
error rate slightly better than random guessing, then adds the weak
classifier to the strong classifier to construct an $m$-layer
network. The weaker classifier is trained by
Algorithm~\ref{alg:nn-rand} (or by Algorithm~\ref{alg:linear-model-l2l2} or Algorithm~\ref{alg:linear-model-linfl1} if $\nn_{m-1}$ are one-layer networks).  The following theorem provides
guarantees for its performance when it is run for 
\[T \defeq \Big\lceil
\frac{16 B^2 \log (n+1)}{\gamma^2}\Big\rceil\] iterations.  The running
time depends on a quantity $C_{m, B, 1/\gamma}$ that is a constant for
any choice of the triple $(m, B, 1/\gamma)$, but with exponential
dependence on $1/\gamma$.

\begin{theorem}
\label{theorem:separable-data-error}
With the above choice of $T$, the BoostNet algorithm achieves:
\begin{enumerate}[(a)]
\item In-sample error: For any $\gamma$-separable
  dataset $\{(x_i, y_i)\}_{i=1}^\numobs$ Algorithm~\ref{alg:nn-fw}
  outputs a neural network $\fhat \in \nn_m$ such that, 
\begin{align*}
y_i \fhat(x_i) \geq \frac{\gamma}{16} \quad \mbox{for every $i\in[n]$,
  $\quad$ with probability at least $1 - \delta$.}
\end{align*}
The time complexity is bounded
  by $\poly(n,d,\log(1/\delta),C_{m,B,1/\gamma})$.
\item Generalization error: Given a data set consisting of $n =
  \poly(1/\epsilon,\log(1/\delta))$ i.i.d. samples from any
  \mbox{$\gamma$-separable} distribution $\mprob$,
  Algorithm~\ref{alg:nn-fw} outputs a network $\fhat\in\nn_m$ such
  that
\begin{align}
\mprob \big[ {\rm sign}(\fhat(x)) \neq y \big] & \leq \epsilon \qquad
\mbox{with probability at least $1-2\delta$.}
\end{align}
Moreover, the time complexity is bounded by
$\poly(d,1/\epsilon,\log(1/\delta), C_{m, B, 1/\gamma})$.
\end{enumerate}
\end{theorem}
\footnotetext{We may choose the hyper-parameters of Algorithm~\ref{alg:nn-rand} by equation~\eqref{eqn:setting-k-s-T}, with the additive error $\epsilon$ defined by $\gamma/((4m+10)LB^m)$. Theorem~\ref{theorem:nn-rand} guarantees that the error bound~\eqref{eqn:alg-nn-fw-weak-learner} holds with high probability.
}

See Appendix~\ref{sec:proof-boosting} for the proof.  The most
technical work is devoted to proving part (a).  The generalization
bound in part (b) follows by combining part (a) with bounds on the
Rademacher complexity of the network class, which then allow us to
translate the in-sample error bound to generalization error in the
usual way. It is worth comparing the BoostNet algorithm with the
general algorithm for agnostic learning. In order to bound the
generalization error by $\epsilon > 0$, the time complexity of
Algorithm~\ref{alg:nn-rand} will be exponential in $1/\epsilon$.

The same learnability results can be established even if the labels are
randomly corrupted. Formally, for every pair $(x,y)$ sampled from a
$\gamma$-separable distribution, suppose that the learning algorithm
actually receives the corrupted pair $(x,\tildey)$, where
\begin{align*}
\tildey =  \begin{cases}
		y & \mbox{with probability $1-\eta$,}\\
		-y & \mbox{with probability $\eta$.}
\end{cases}
\end{align*}
Here the parameter $\eta \in [0, \frac{1}{2})$ corresponds to the
noise level. Since the labels are flipped, the BoostNet algorithm
cannot be directly applied. However, we can use the improper
learning algorithm of \cite{zhang2015ell_1} to learn an
improper classifier $\hhat$ such that $\hhat(x)=y$ with high
probability, and then apply the BoostNet algorithm taking
$(x,\hhat(x))$ as input.  Doing so yields the following guarantee:

\begin{corollary}\label{coro:flip-label}
Assume that $q = 2$ and $\eta < 1/2$. For any constant $(m,B)$,
consider the neural network class $\nn_m$ activated by $\sigma(x)
\defeq {\rm erf}(x)$\footnote{The erf function can be replaced by any
function $\sigma$ satisfying polynomial expansion $\sigma(x)
  = \sum_{j=0}^\infty \beta_jx^j$, such that $\sum_{j=0}^\infty
  2^j \beta_j^{2} \lambda^{2j} < +\infty$ for any finite
  $\lambda\in\R^+$.}.  Given a random dataset of size $n =
\poly(1/\epsilon, 1/\delta)$ for any $\gamma$-separable distribution,
there is a $\poly(d,1/\epsilon,1/\delta)$-time algorithm that outputs
a network $\fhat\in \nn_m$ such that 
\begin{align*}
\mprob({\rm sign}(\fhat(x)) \neq y) \leq \epsilon \qquad 
\mbox{with probability at least $1-\delta$.}
\end{align*}
\end{corollary}
\noindent See Appendix~\ref{sec:proof-flip-label} for the proof.


\subsection{Hardness result for $\gamma$-separable problems}

Finally, we present a hardness result showing that the dependence on
$1/\gamma$ is hard to improve. Our proof relies on the hardness of
standard (nonagnostic) PAC learning of the intersection of halfspaces
given in \cite{klivans2006cryptographic}. More
precisely, consider the family of halfspace indicator functions
mapping $\mathcal{X} = \{-1,1\}^d$ to $\{-1,1\}$ given by
\begin{align*}
H = \{ x\to {\rm sign}(w^T x - b - 1/2): x\in\{-1,1\}^d,~b\in \N,
~w\in \N^d,~|b| + \lones{w} \leq \poly(d) \}.
\end{align*}
Given a $T$-tuple of functions $\{h_1, \ldots, h_T \}$ belonging to
$H$, we define the intersection function
\begin{align*}
	h(x) = \begin{cases} 1 & \mbox{if $h_1(x) = \dots = h_T(x) =
            1$},\\ 
-1 & \mbox{otherwise},
        \end{cases}
\end{align*}
which represents the intersection of $T$ half-spaces.  Letting $H_T$
denote the set of all such functions, for any distribution on
$\mathcal{X}$, we want an algorithm taking a sequence of $(x,h^*(x))$
as input where $x$ is a sample from $\mathcal{X}$ and $h^*\in H_T$. It
should output a function $\hhat$ such that $\mprob(\hhat(x) \neq
h^*(x)) \leq \epsilon$ with probability at least $1-\delta$.  If there
is such an algorithm whose sample complexity and time complexity scale
as $\poly(d,1/\epsilon,1/\delta)$, then we say that $H_T$ is
efficiently learnable. \cite{klivans2006cryptographic} show that if $T =
\Theta(d^{\rho})$ then $H_T$ is not efficiently learnable under a
certain cryptographic assumption. This hardness statement implies the
hardness of learning neural networks with separable data. 
\begin{proposition}
\label{theorem:hardness-neuralnet-margin}
Assume $H_T$ is not efficiently learnable for $T = \Theta(d^{\rho})$.
Consider the class of two-layer neural networks $\nn_2$ activated by
the piecewise linear function $\sigma(x) \defeq
\min\{1,\max\{-1,x\}\}$ or the ReLU function $\sigma(x) \defeq
\max\{0,x\}$, and with the norm constraint $B = 2$.  Consider any
algorithm such that when applied to any $\gamma$-separable data
distribution, it is guaranteed to output a neural network $\fhat$
satisfying $\mprob({\rm sign}(\fhat(x))\neq y) \leq
\epsilon$
\mbox{with probability at least~$1-\delta$.}
Then it cannot run in $\poly(d,1/\epsilon, 1/\delta, 1/\gamma)$-time.

\end{proposition}
\noindent
See Appendix~\ref{sec:proof-hardness-neuralnet-margin} for the proof.


\section{Simulation}
\label{sec:simulation}

In this section, we compare the BoostNet algorithm with the classical
backpropagation method for training two-layer neural networks. The
goal is to learn parity functions from noisy data --- a challenging
problem in computational learning theory~\cite[see, e.g.][]{blum2003noise}.  We
construct a synthetic dataset with $n = 50,000$ points. Each point
$(x,y)$ is generated as follows: first, the vector $x$ is uniformly
drawn from $\{-1,1\}^{d}$ and concatenated with a constant $1$ as the
$(d+1)$-th coordinate. The label is generated as follows: for some
unknown subset of $p$ indices $1 \leq i_1 < \dots < i_p \leq d$, we
set
\begin{align*}
y = \begin{cases} x_{i_1}x_{i_2}\dots x_{i_p} & \mbox{with probability
    0.9} \,,\\ 
- x_{i_1}x_{i_2}\dots x_{i_p} & \mbox{with probability
    0.1} \, .
\end{cases}
\end{align*}
The goal is to learn a function $f:\R^{d+1}\to \R$ such that ${\rm
  sign}(f(x))$ predicts the value of $y$. The optimal rate is achieved
by the parity function $f(x) = x_{i_1}x_{i_2}\dots x_{i_p}$, in which
case the prediction error is $0.1$. If the parity degree $p > 1$, the
optimal rate cannot be achieved by any linear classifier.

\begin{figure}
\begin{tabular}{ccc}
\includegraphics[height = 0.3\textwidth]{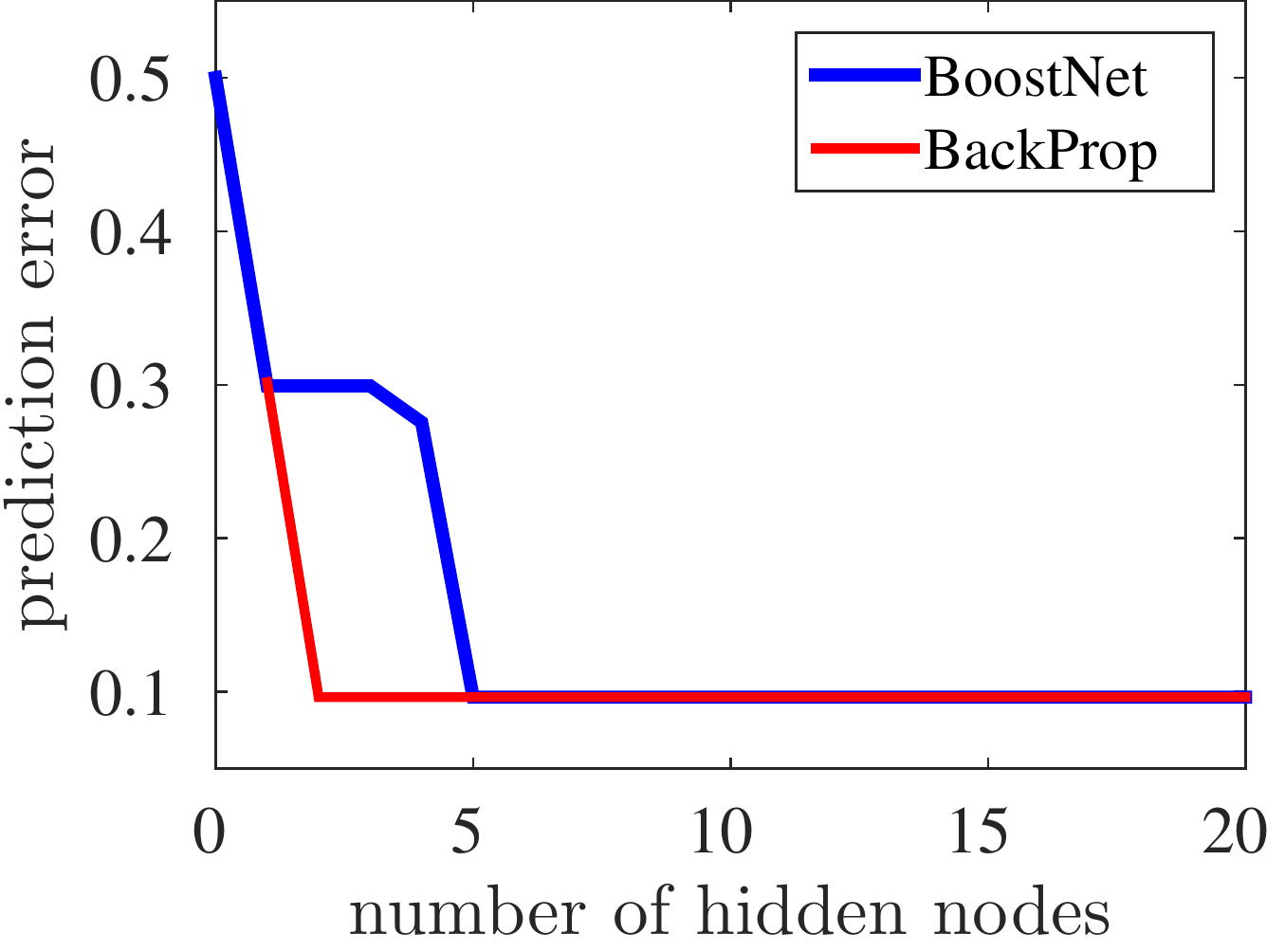} &~~~~&
\includegraphics[height = 0.3\textwidth]{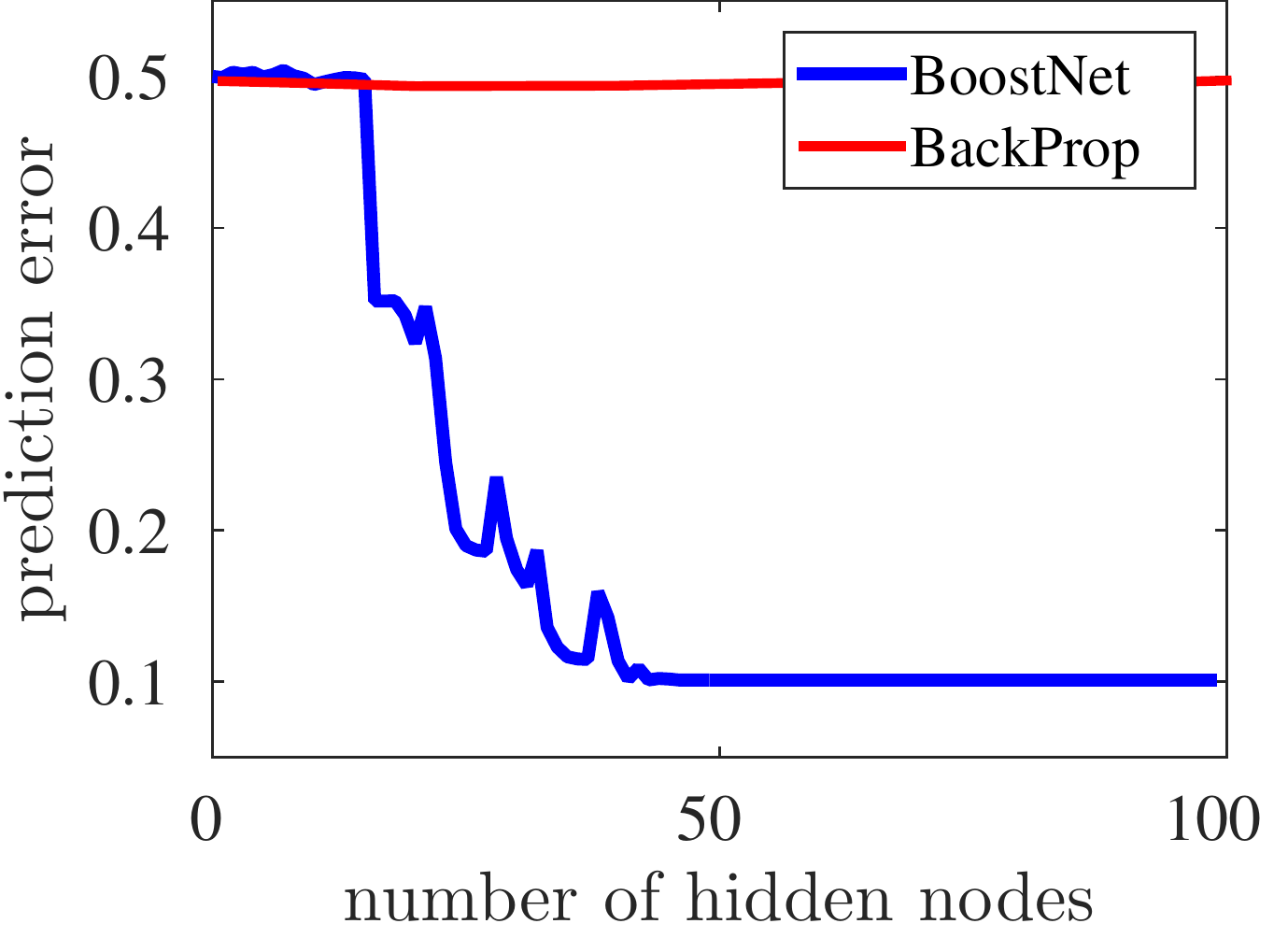} \\
(a) parity degree $p = 2$ && (b) parity degree $p = 5$
\end{tabular}
\label{fig:plot-parity}
\caption{Performance of BoostNet and 
BackProp on the problem of learning parity function with noise.}
\end{figure}

Choose $d=50$ and $p\in\{2,5\}$. The activation function is chosen as
$\sigma(x) \defeq \tanh(x)$. The training set, the validation set and
the test set contain respectively 25K, 5K and 20K points. To train a
two-layer BoostNet, we choose the hyper-parameter $B=10$, and select
Algorithm~\ref{alg:nn-rand} as the subroutine to train weak
classifiers with hyper-parameters $(k,T) = (10,1)$. To train a
classical two-layer neural network, we use the random initialization scheme of \cite{Nguyen1990} and the backpropagation algorithm of \cite{moller1993scaled}. For both methods,
the algorithm is executed for ten independent rounds to select the
best solution.

Figure~\ref{fig:plot-parity} compares the prediction errors of
BoostNet and backpropagation. Both methods generate the same two-layer
network architecture, so that we compare with respect to the number of
hidden nodes. Note that BoostNet constructs hidden nodes incrementally
while NeuralNet trains a predefined number of
neurons. Figure~\ref{fig:plot-parity} shows that both algorithms learn
the degree-2 parity function with a few hidden nodes. In contrast,
BoostNet learns the degree-5 parity function with less than 50 hidden
nodes, while NeuralNet's performance is no better than random
guessing. This suggests that the BoostNet algorithm is less likely to
be trapped in a bad local optimum in this setting.

\section{Conclusion}

In this paper, we have proposed algorithms to learn halfspaces and
neural networks with non-convex loss functions. We demonstrate that
the time complexity is polynomial in the input dimension and in the
sample size but exponential in the excess risk. A hardness result
relating to the necessity of the exponential dependence is also
presented. The algorithms perform randomized initialization followed
by optimization steps. This idea coincides with the heuristics that
are widely used in practice, but the theoretical analysis suggests
that a careful treatment of the initialization step is necessary. We
proposed the BoostNet algorithm, and showed that it can be used to
learn a neural network in polynomial time when the data are separable
with a constant margin. We suspect that the theoretical results of
this paper are likely conservative, in that in application to real
data, the algorithms can be much more efficient than the bounds might
suggest.


\subsection*{Acknowledgements:}  MW and YZ were partially supported by
NSF grant CIF-31712-23800 from the National Science Foundation,
AFOSR-FA9550-14-1-0016 grant from the Air Force Office of Scientific
Research, ONR MURI grant N00014-11-1-0688 from the Office of Naval
Research. MJ and YZ were partially supported by the U.S.ARL
and the U.S.ARO under contract/grant number W911NF-11-1-0391.
We thank Sivaraman Balakrishnan for helpful comments on an
earlier draft.

\bibliographystyle{abbrvnat} 
\bibliography{bib}

\begin{thebibliography}{39}
\providecommand{\natexlab}[1]{#1}
\providecommand{\url}[1]{\texttt{#1}}
\expandafter\ifx\csname urlstyle\endcsname\relax
  \providecommand{\doi}[1]{doi: #1}\else
  \providecommand{\doi}{doi: \begingroup \urlstyle{rm}\Url}\fi

\bibitem[Arora et~al.(2013)Arora, Bhaskara, Ge, and Ma]{arora2013provable}
S.~Arora, A.~Bhaskara, R.~Ge, and T.~Ma.
\newblock Provable bounds for learning some deep representations.
\newblock \emph{arXiv:1310.6343}, 2013.

\bibitem[Awasthi et~al.(2014)Awasthi, Balcan, and Long]{awasthi2014power}
P.~Awasthi, M.~F. Balcan, and P.~M. Long.
\newblock The power of localization for efficiently learning linear separators
  with noise.
\newblock In \emph{Proceedings of the 46th Annual ACM Symposium on Theory of
  Computing}, pages 449--458. ACM, 2014.

\bibitem[Awasthi et~al.(2015)Awasthi, Balcan, Haghtalab, and
  Urner]{awasthi2015efficient}
P.~Awasthi, M.-F. Balcan, N.~Haghtalab, and R.~Urner.
\newblock Efficient learning of linear separators under bounded noise.
\newblock \emph{arXiv:1503.03594}, 2015.

\bibitem[Barron(1993)]{barron1993universal}
A.~R. Barron.
\newblock Universal approximation bounds for superpositions of a sigmoidal
  function.
\newblock \emph{IEEE Transactions on Information Theory}, 39\penalty0
  (3):\penalty0 930--945, 1993.

\bibitem[Bartlett(1998)]{bartlett1998sample}
P.~L. Bartlett.
\newblock The sample complexity of pattern classification with neural networks:
  the size of the weights is more important than the size of the network.
\newblock \emph{Information Theory, IEEE Transactions on}, 44\penalty0
  (2):\penalty0 525--536, 1998.

\bibitem[Bartlett and Mendelson(2003)]{bartlett2003rademacher}
P.~L. Bartlett and S.~Mendelson.
\newblock Rademacher and gaussian complexities: Risk bounds and structural
  results.
\newblock \emph{The Journal of Machine Learning Research}, 3:\penalty0
  463--482, 2003.

\bibitem[Ben-David and Simon(2001)]{simon2001efficient}
S.~Ben-David and H.~U. Simon.
\newblock Efficient learning of linear perceptrons.
\newblock In \emph{Advances in Neural Information Processing Systems},
  volume~13, page 189. MIT Press, 2001.

\bibitem[Birnbaum and Shwartz(2012)]{birnbaum2012learning}
A.~Birnbaum and S.~S. Shwartz.
\newblock Learning halfspaces with the zero-one loss: time-accuracy tradeoffs.
\newblock In \emph{Advances in Neural Information Processing Systems},
  volume~24, pages 926--934, 2012.

\bibitem[Blum and Rivest(1992)]{blum1992training}
A.~Blum and R.~L. Rivest.
\newblock Training a 3-node neural network is {NP}-complete.
\newblock \emph{Neural Networks}, 5\penalty0 (1):\penalty0 117--127, 1992.

\bibitem[Blum et~al.(1998)Blum, Frieze, Kannan, and
  Vempala]{blum1998polynomial}
A.~Blum, A.~Frieze, R.~Kannan, and S.~Vempala.
\newblock A polynomial-time algorithm for learning noisy linear threshold
  functions.
\newblock \emph{Algorithmica}, 22\penalty0 (1-2):\penalty0 35--52, 1998.

\bibitem[Blum et~al.(2003)Blum, Kalai, and Wasserman]{blum2003noise}
A.~Blum, A.~Kalai, and H.~Wasserman.
\newblock Noise-tolerant learning, the parity problem, and the statistical
  query model.
\newblock \emph{Journal of the ACM}, 50\penalty0 (4):\penalty0 506--519, 2003.

\bibitem[Daniely et~al.(2014)Daniely, Linial, and
  Shalev-Shwartz]{daniely2014average}
A.~Daniely, N.~Linial, and S.~Shalev-Shwartz.
\newblock From average case complexity to improper learning complexity.
\newblock In \emph{Proceedings of the 46th Annual ACM Symposium on Theory of
  Computing}, pages 441--448. ACM, 2014.

\bibitem[Dasgupta and Gupta(1999)]{dasgupta1999elementary}
S.~Dasgupta and A.~Gupta.
\newblock An elementary proof of the {J}ohnson-{L}indenstrauss lemma.
\newblock \emph{International Computer Science Institute, Technical Report},
  pages 99--006, 1999.

\bibitem[Freund and Schapire(1997)]{freund1997decision}
Y.~Freund and R.~E. Schapire.
\newblock A decision-theoretic generalization of on-line learning and an
  application to boosting.
\newblock \emph{Journal of Computer and System Sciences}, 55\penalty0
  (1):\penalty0 119--139, 1997.

\bibitem[Guruswami and Raghavendra(2009)]{guruswami2009hardness}
V.~Guruswami and P.~Raghavendra.
\newblock Hardness of learning halfspaces with noise.
\newblock \emph{SIAM Journal on Computing}, 39\penalty0 (2):\penalty0 742--765,
  2009.

\bibitem[Hush(1999)]{hush1999training}
D.~R. Hush.
\newblock Training a sigmoidal node is hard.
\newblock \emph{Neural Computation}, 11\penalty0 (5):\penalty0 1249--1260,
  1999.

\bibitem[Janzamin et~al.(2015)Janzamin, Sedghi, and
  Anandkumar]{janzamin2015generalization}
M.~Janzamin, H.~Sedghi, and A.~Anandkumar.
\newblock Generalization bounds for neural networks through tensor
  factorization.
\newblock \emph{arXiv:1506.08473}, 2015.

\bibitem[Kakade and Tewari(2008)]{Kakade2008lecture}
S.~Kakade and A.~Tewari.
\newblock Lecture note: Rademacher composition and linear prediction.
\newblock 2008.

\bibitem[Kakade et~al.(2009)Kakade, Sridharan, and
  Tewari]{kakade2009complexity}
S.~M. Kakade, K.~Sridharan, and A.~Tewari.
\newblock On the complexity of linear prediction: Risk bounds, margin bounds,
  and regularization.
\newblock In \emph{Advances in Neural Information Processing Systems},
  volume~21, pages 793--800, 2009.

\bibitem[Kalai et~al.(2008)Kalai, Klivans, Mansour, and
  Servedio]{kalai2008agnostically}
A.~T. Kalai, A.~R. Klivans, Y.~Mansour, and R.~A. Servedio.
\newblock Agnostically learning halfspaces.
\newblock \emph{SIAM Journal on Computing}, 37\penalty0 (6):\penalty0
  1777--1805, 2008.

\bibitem[Klivans and Kothari(2014)]{klivans2014embedding}
A.~Klivans and P.~Kothari.
\newblock Embedding hard learning problems into gaussian space.
\newblock \emph{Approximation, Randomization, and Combinatorial Optimization.
  Algorithms and Techniques}, 28:\penalty0 793--809, 2014.

\bibitem[Klivans et~al.(2006)Klivans, Sherstov,
  et~al.]{klivans2006cryptographic}
A.~R. Klivans, A.~Sherstov, et~al.
\newblock Cryptographic hardness for learning intersections of halfspaces.
\newblock In \emph{47th Annual IEEE Symposium on Foundations of Computer
  Science}, pages 553--562. IEEE, 2006.

\bibitem[Klivans et~al.(2009)Klivans, Long, and Servedio]{klivans2009learning}
A.~R. Klivans, P.~M. Long, and R.~A. Servedio.
\newblock Learning halfspaces with malicious noise.
\newblock \emph{The Journal of Machine Learning Research}, 10:\penalty0
  2715--2740, 2009.

\bibitem[Koltchinskii and Panchenko(2002)]{koltchinskii2002empirical}
V.~Koltchinskii and D.~Panchenko.
\newblock Empirical margin distributions and bounding the generalization error
  of combined classifiers.
\newblock \emph{Annals of Statistics}, pages 1--50, 2002.

\bibitem[Ledoux and Talagrand(2013)]{Ledoux2013probability}
M.~Ledoux and M.~Talagrand.
\newblock \emph{Probability in Banach Spaces: isoperimetry and processes},
  volume~23.
\newblock Springer Science \& Business Media, 2013.

\bibitem[Livni et~al.(2014)Livni, Shalev-Shwartz, and
  Shamir]{livni2014computational}
R.~Livni, S.~Shalev-Shwartz, and O.~Shamir.
\newblock On the computational efficiency of training neural networks.
\newblock In \emph{Advances in Neural Information Processing Systems},
  volume~26, pages 855--863, 2014.

\bibitem[M{\o}ller(1993)]{moller1993scaled}
M.~F. M{\o}ller.
\newblock A scaled conjugate gradient algorithm for fast supervised learning.
\newblock \emph{Neural networks}, 6\penalty0 (4):\penalty0 525--533, 1993.

\bibitem[Neyshabur et~al.(2015)Neyshabur, Tomioka, and
  Srebro]{neyshabur2015norm}
B.~Neyshabur, R.~Tomioka, and N.~Srebro.
\newblock Norm-based capacity control in neural networks.
\newblock \emph{arXiv preprint arXiv:1503.00036}, 2015.

\bibitem[Nguyen and Widrow(1990)]{Nguyen1990}
D.~Nguyen and B.~Widrow.
\newblock Improving the learning speed of 2-layer neural networks by choosing
  initial values of the adaptive weights.
\newblock In \emph{International Joint Conference on Neural Networks}, pages
  21--26, 1990.

\bibitem[Pisier(1980)]{pisier1980remarques}
G.~Pisier.
\newblock Remarques sur un r{\'e}sultat non publi{\'e} de {B}. {M}aurey.
\newblock \emph{S{\'e}minaire Analyse Fonctionnelle}, pages 1--12, 1980.

\bibitem[Rosenblatt(1958)]{rosenblatt1958perceptron}
F.~Rosenblatt.
\newblock The perceptron: a probabilistic model for information storage and
  organization in the brain.
\newblock \emph{Psychological Review}, 65\penalty0 (6):\penalty0 386, 1958.

\bibitem[Schapire and Singer(1999)]{schapire1999improved}
R.~E. Schapire and Y.~Singer.
\newblock Improved boosting algorithms using confidence-rated predictions.
\newblock \emph{Machine Learning}, 37\penalty0 (3):\penalty0 297--336, 1999.

\bibitem[Sedghi and Anandkumar(2014)]{sedghi2014provable}
H.~Sedghi and A.~Anandkumar.
\newblock Provable methods for training neural networks with sparse
  connectivity.
\newblock \emph{arXiv:1412.2693}, 2014.

\bibitem[Servedio and Valiant(2001)]{servedio2001efficient}
R.~Servedio and L.~Valiant.
\newblock Efficient algorithms in computational learning theory.
\newblock \emph{Harvard University, Cambridge, MA}, 2001.

\bibitem[Shalev-Shwartz and Singer(2010)]{shalev2010equivalence}
S.~Shalev-Shwartz and Y.~Singer.
\newblock On the equivalence of weak learnability and linear separability: New
  relaxations and efficient boosting algorithms.
\newblock \emph{Machine Learning}, 80\penalty0 (2-3):\penalty0 141--163, 2010.

\bibitem[Shalev-Shwartz et~al.(2011)Shalev-Shwartz, Shamir, and
  Sridharan]{shalev2011learning}
S.~Shalev-Shwartz, O.~Shamir, and K.~Sridharan.
\newblock Learning kernel-based halfspaces with the 0-1 loss.
\newblock \emph{SIAM Journal on Computing}, 40\penalty0 (6):\penalty0
  1623--1646, 2011.

\bibitem[Thom and Palm(2013)]{thom2013sparse}
M.~Thom and G.~Palm.
\newblock Sparse activity and sparse connectivity in supervised learning.
\newblock \emph{The Journal of Machine Learning Research}, 14\penalty0
  (1):\penalty0 1091--1143, 2013.

\bibitem[Vapnik(1998)]{vapnik1998statistical}
V.~N. Vapnik.
\newblock \emph{Statistical learning theory}, volume~1.
\newblock Wiley New York, 1998.

\bibitem[Zhang et~al.(2015)Zhang, Lee, and Jordan]{zhang2015ell_1}
Y.~Zhang, J.~D. Lee, and M.~I. Jordan.
\newblock $\ell_1$-regularized neural networks are improperly learnable in
  polynomial time.
\newblock \emph{arXiv:1510.03528}, 2015.

\end{thebibliography}

\appendix

\section{Proof of Lemma~\ref{lemma:rademacher-generalization}}
\label{sec:proof-rademacher-generalization}

The following inequality always holds:
\begin{align*}
\sup_{f\in \mathcal{F}} |G(f)-\ell(f)|  \leq \max\Big\{ \sup_{f\in \mathcal{F}} \{G(f)-\ell(f)\}, \sup_{f'\in \mathcal{F}} \{\ell(f') - G(f') \} \Big\}.
\end{align*}
Since $\mathcal{F}$ contains the constant zero function, both $\sup_{f\in \mathcal{F}} \{G(f)-\ell(f)\}$ and $\sup_{f'\in \mathcal{F}} \{\ell(f') - G(f') \}$
are non-negative, which implies
\begin{align*}
\sup_{f\in \mathcal{F}} |G(f)-\ell(f)| \leq \sup_{f\in \mathcal{F}} \{G(f)-\ell(f)\} + \sup_{f'\in \mathcal{F}} \{\ell(f') - G(f') \}.
\end{align*}
To establish Lemma~\ref{lemma:rademacher-generalization}, it suffices to prove: 
\[
	\E \Big[\sup_{f\in \mathcal{F}} \{G(f)-\ell(f)\}\Big] \leq 2 L R_k(\mathcal{F})
	\quad \mbox{and} \quad \E \Big[\sup_{f'\in \mathcal{F}} \{\ell(f') - G(f')\}\Big] \leq 2 L R_k(\mathcal{F})
\]
For the rest of the proof, we will establish the first upper bound. The second  bound can be established through an identical series of steps.

The inequality $\E [\sup_{f\in \mathcal{F}} \{G(f)-\ell(f)\}] \leq 2 L R_k(\mathcal{F})$ follows as a consequence
of classical symmetrization techniques \cite[e.g.][]{bartlett2003rademacher} and the Talagrand-Ledoux
concentration \cite[e.g.][Corollary
3.17]{Ledoux2013probability}. However, so as to keep the paper
self-contained, we provide a detailed proof here.  By the definitions
of $\ell(f)$ and $G(f)$, we have
\begin{align*}
\E \Big[\sup_{f\in \mathcal{F}} \Big\{G(f)-\ell(f)\Big\}\Big] =
\E\Big[\sup_{f\in \mathcal{F}}  \Big\{\frac{1}{k} \sum_{j=1}^k
  h(-y'_jf(x'_j)) - \E\Big[ \frac{1}{k} \sum_{j=1}^k
    h(-y''_jf(x''_j)) \Big]\Big\} \Big],
\end{align*}
where $(x''_j,y''_j)$ is an i.i.d.~copy of $(x'_j,y'_j)$. Applying
Jensen's inequality yields
\begin{align}
\E\Big[\sup_{f\in \mathcal{F}} \Big\{G(f)-\ell(f)\Big\}\Big] & \leq
\E\Big[\sup_{f\in \mathcal{F}} \Big\{ \frac{1}{k} \sum_{j=1}^k
  h(-y'_jf(x'_j)) - h(-y''_jf(x''_j)) \Big\}\Big] \nonumber \\
&
  = \E \Big[\sup_{f\in \mathcal{F}} \Big\{ \frac{1}{k} \sum_{j=1}^k \varepsilon_j(h(-y'_jf(x'_j))
  - h(-y''_jf(x''_j))) \Big\}\Big]
\nonumber \\ &\leq \E\Big[\sup_{f\in \mathcal{F}} \Big\{ \frac{1}{k}
  \sum_{j=1}^k \varepsilon_j h(-y'_jf(x'_j))
    + \sup_{f\in \mathcal{F}}  \frac{1}{k} \sum_{j=1}^k \varepsilon_j
    h(-y''_jf(x''_j)) \Big\}\Big] \nonumber \\
 & =
  2\E\Big[\sup_{f\in \mathcal{F}} \Big\{ \frac{1}{k} \sum_{j=1}^k \varepsilon_j
  h(-y'_jf(x'_j))\Big\}\Big].\label{eqn:gf-to-2rad}
\end{align}
We need to bound the right-hand side using the Rademacher complexity
of the function class~$\mathcal{F}$, and we use an argument following
the lecture notes of \cite{Kakade2008lecture}.
Introducing the shorthand notation $\varphi_j(x) \defeq h(-y'_j x)$,
the $L$-Lipschitz continuity of $\varphi_j$ implies that
\begin{align*}
\E\Big[\sup_{f\in \mathcal{F}}  \sum_{j=1}^k \varepsilon_j
  \varphi_j(f(x'_j))\Big]
  &= \E\Big[\sup_{f,f'\in \mathcal{F}} \Big\{ \frac{\varphi_1(f(x'_1))
  - \varphi_1(f'(x'_1))}{2}
  + \sum_{j=2}^k \varepsilon_j \frac{\varphi_j(f(x'_j))
  + \varphi_j(f'(x'_j))}{2}\Big\}\Big] \\
& \leq \E\Big[\sup_{f,f'\in \mathcal{F}} \Big\{ \frac{L|f(x'_1) -
    f'(x'_1)|}{2}
    + \sum_{j=2}^k \varepsilon_j \frac{\varphi_j(f(x'_j))
    + \varphi_j(f'(x'_j))}{2}\Big\}\Big] \\
& = \E\Big[\sup_{f,f'\in \mathcal{F}}  \Big\{\frac{L f(x'_1) - L
    f'(x'_1)}{2} + \sum_{j=2}^k \varepsilon_j \frac{\varphi_j(f(x'_j))
    + \varphi_j(f'(x'_j))}{2}\Big\}\Big].
\end{align*}
Applying Jensen's inequality implies that the right-hand side is
bounded by
\begin{align*}
{\rm RHS} & \leq \frac{1}{2}\E\Big[\sup_{f\in \mathcal{F}} \Big\{ L
  f(x'_1) + \sum_{j=2}^k \varepsilon_j \varphi_j(f(x'_j)) \Big\}
  + \sup_{f'\in \mathcal{F}}  \Big\{-L f(x'_1)
  + \sum_{j=2}^k \varepsilon_j \varphi_j(f'(x'_j))\Big\} \Big] \\
& = \E\Big[\sup_{f\in \mathcal{F}} \Big\{ \varepsilon_1 L f(x'_1)
  + \sum_{j=2}^k \varepsilon_j \varphi_j(f(x'_j)) \Big\}\Big].
\end{align*}
By repeating this argument for $j = 2,3,\dots,k$, we obtain
\begin{align}
\E\Big[\sup_{f\in \mathcal{F}}  \sum_{j=1}^k \varepsilon_j
  \varphi_j(f(x'_j))\Big] \leq
  L \E\Big[\sup_{f\in \mathcal{F}}  \sum_{j=1}^k \varepsilon_j
  f(x'_j)\Big]. \label{eqn:radphi-to-radabs}
\end{align}
Combining inequalities~\eqref{eqn:gf-to-2rad}
and~\eqref{eqn:radphi-to-radabs}, we have the desired bound.


\section{Proof of Theorem~\ref{alg:linear-model-l2l2}}
\label{sec:proof-theorem-l2l2}

Let us study a single iteration of
Algorithm~\ref{alg:linear-model-l2l2}.  Recall that $u$ is uniformly
sampled from the unit sphere of $\R^d$. Alternatively, it can be
viewed as sampled by the following procedure: first draw a random
$s$-dimensional subspace of $\R^d$, then draw $u$ uniformly from the
unit sphere of $S$. Consider the $n+2$ fixed points
$\{0,\wstar,x_1,\dots,x_n\}$. Let $\phi(w) := r\Pi_{S}(w)$ be an
operator that projects the vector $w\in \R^d$ to the subspace $S$ and
scale it by $r$. If we choose $s\geq \frac{12 \log (n+2)}{\epsilon^2}$
and $r\defeq \sqrt{d/s}$, then Lemma~\ref{lemma:jl} implies that with
probability at least $1/(n+2)$, we are guaranteed that
\begin{align}
|\ltwos{\phi(\wstar)}^2
  |- \ltwos{\wstar}^2| \leq \epsilon, \qquad \ltwos{\phi(x_i)}^2
  |- \ltwos{x_i}^2| \leq \epsilon \quad \mbox{and} \quad \nonumber\\
\label{eqn:inner-product-dist}
| \ltwos{\phi(\wstar) - \phi(x_i) }^2 - \ltwos{\wstar - x_i}^2 | \leq
  |4\epsilon \quad \mbox{for any $i\in[n]$}.
\end{align}
Consequently, we have
\begin{align}
\label{eqn:inner-product-approx}
|\langle \phi(\wstar), \phi(x_i)\rangle - \langle \wstar, x_i\rangle|
\leq 3\epsilon \quad \mbox{for any $i\in[n]$}.
\end{align}
Assume that the approximation bounds~\eqref{eqn:inner-product-dist}
and~\eqref{eqn:inner-product-approx} hold. Then using the
$L$-Lipschitz continuity of $h$, we have
\begin{align}
\label{eqn:approx-by-sup}
\ell(\wstar) & = \sum_{i=1}^n \alpha_i h(-y_i \langle \wstar,
x_i\rangle) \geq - 3\epsilon L + \sum_{j=1}^k \alpha_i h(-y_i \langle
\phi(\wstar), \phi(x_i)\rangle).
\end{align}
Recall that $u$ is uniformly drawn from the unit sphere of $S$;
therefore, the angle between $u$ and $\phi(\wstar)$ is at most
$\epsilon$ with probability at least $(\epsilon/\pi)^{s-1}$.  By
inequality~\eqref{eqn:inner-product-dist}, the norm of $\phi(\wstar)$
is in $[1-\epsilon,1+\epsilon]$. Hence, whenever the angle bound
holds, the distance between $u$ and $\phi(\wstar)$ is bounded by
$2\epsilon$. Combining this bound with
inequality~\eqref{eqn:approx-by-sup}, we find that
\begin{align*}
\ell(\wstar) \geq - 3\epsilon L + \sum_{i=1}^n \alpha_i h(-y_i \langle
u + (\phi(\wstar) - u), \phi(x_i)\rangle).
\end{align*}
Note that we have the inequality
\begin{align*}
|\langle \phi(\wstar) - u, \phi(x_i)\rangle| \leq \ltwos{\phi(\wstar)
- u}\ltwos{\phi(x_i)} \leq 2\epsilon (1+\epsilon)\leq 3\epsilon.
\end{align*}
Combined with the $L$-Lipschitz condition, we obtain the lower bound
\begin{align}
\ell(\wstar) \geq - 6\epsilon L + \sum_{j=1}^k \alpha_i h( -
y_i\langle u, \phi(x_i)\rangle) = \ell(ru) - 6\epsilon
L.\label{eqn:approx-solution-bound}
\end{align}
This bound holds with probability at least
$\frac{(\epsilon/\pi)^{s-1}}{2n+4}$. Thus, by repeating the procedure
for a total of \mbox{$T \geq (2n+4)
(\pi/\epsilon)^{s-1}\log(1/\delta)$ iterations,} then the best
solution satisfies inequality~\eqref{eqn:approx-solution-bound} with
probability at least $1-\delta$. The time complexity is obtained by
plugging in the stated choices of $(s,T)$.


\section{Proof of Theorem~\ref{theorem:bound-linfl1}}
\label{sec:proof-bound-linfl1}

Let us study a single iteration of
Algorithm~\ref{alg:linear-model-linfl1}. Conditioning on any
$\{(x'_j,y'_j)\}_{j=1}^k$, define the function \mbox{$G(w) \defeq
\frac{1}{k}\sum_{j=1}^k h(-y'_j \langle w, x'_j\rangle)$.}  Since $h$
is $L$-Lipschitz, we have
\begin{align*}
G(v) - G(\wstar) = \frac{1}{k}\sum_{j=1}^k h(-y'_j \langle v,
x'_j\rangle) - h(-y'_j \langle \wstar, x'_j\rangle) & \leq \frac{L}{k}
\sum_{j=1}^k |\langle v - \wstar, x'_j\rangle| \\ 
& \leq \frac{L}{\sqrt{k}} \Big( \sum_{j=1}^k (\langle v - \wstar,
x'_j\rangle)^2 \Big)^{1/2},
\end{align*}
where the final step follows from the Cauchy-Schwarz inequality. Let
$X' \in \R^{k\times d}$ be a design matrix whose $j$-th row is equal
to $x'_j$. Using the above bound, for any $u\in [-1,1]^k$, we have
\begin{align}
G(v) - G(\wstar)
& \leq \frac{L}{\sqrt{k}} \ltwos{X'(v-\wstar)} \leq \frac{L}{\sqrt{k}}
(\ltwos{X'v-u} + \ltwos{X'\wstar-u})\nonumber \\
& \leq \frac{2L}{\sqrt{k}} \ltwos{X'\wstar-u},\label{eqn:G-diff}
\end{align}
where the last step uses the fact that the vector $v$ minimizes
$\ltwos{X'w-u}$ over the set of vectors with $\norms{w}_p \leq 1$. The
right-hand side of inequality~\eqref{eqn:G-diff} is independent of
$v$. Indeed, since $X'\wstar \in [-1,1]^k$ and $u$ is uniformly
sampled from this cube, the probability that $\linftys{X'\wstar-u}
\leq \epsilon/2$ is at least $(\epsilon/4)^k$. This bound leads to
$\ltwos{X'\wstar-u} \leq \epsilon\sqrt{k}/2$. Thus, with probability
at least $(\epsilon/4)^k$, we have $G(v) - G(\wstar) \leq \epsilon L$.

Conditioned on the above bound holding,
Lemma~\ref{lemma:rademacher-generalization} implies that
\begin{align*}
\E \Big[ \sup_{\norms{w}_p\leq 1} \big| \ell(w) - G(w) \big|\Big] \leq
4LR_k(\mathcal{F}),
\end{align*}
where $R_k(\mathcal{F})$ is the $k$-sample Rademacher complexity of
the function class $\mathcal{F} \defeq \{f: x\to \langle w,x\rangle \,
\mid \, \norms{w}_p\leq 1, \norms{x}_q\leq 1\}$.  Markov's inequality
implies that $\sup_{\norms{w}_p\leq 1} \big| \ell(w) - G(w) \big| \leq
5LR_k(\mathcal{F})$ with probability at least $1/5$. This event only
depends on the choice of $\{(x'_j,y'_j)\}$, and conditioned on it
holding, we have
\begin{align*}
\ell(v) & \leq G(v) + 5LR_k(\mathcal{F}) = G(\wstar) + (G(v) -
G(\wstar)) + 5LR_k(\mathcal{F}) \\ 
& \leq \ell(\fstar) + (G(v) - G(\wstar)) + 10LR_k(\mathcal{F}).
\end{align*}
It is known \cite[e.g.][]{kakade2009complexity} that the
Rademacher complexity is upper bounded as
\begin{align*}
R_k(\mathcal{F}) & \leq
\begin{cases} \sqrt{\frac{2\log
    d}{k}} & \mbox{if } p = 1 \\
\sqrt{\frac{q-1}{k}} & \mbox{if $p > 1$, where $1/q + 1/p = 1$.}
\end{cases}
\end{align*}
Thus, given the choice of $k$ from
equation~\eqref{eqn:convex-alg-choose-k} and plugging in the bound on
$G(v) - G(\wstar)$, we have $\ell(v) \leq \ell(\wstar) + 11 \epsilon L$
with probability at least $(\epsilon/4)^k/5$.  Repeating this
procedure for a total of \mbox{$T \geq 5(4/\epsilon)^k\log(1/\delta)$}
times guarantees that the upper bound holds with probability at least
$1-\delta$. Finally, the claimed time complexity is obtained by
plugging in the stated choices of $k$ and $T$.


\section{Proof of Proposition~\ref{theorem:piecewise-linear-hard}}
\label{sec:proof-piecewise-linear-hard}

We prove the lower bound based on a data set with $\ltwos{x_i} \leq 1$
and $y_i = -1$ for each $i\in [n]$. With this set-up, consider the
following problem:
\begin{align}
\label{eqn:hardnes-define-l}
 \mbox{minimize $\ell(w) $ s.t. $\ltwos{w}\leq 1$} \quad
 \mbox{where}\quad \ell(w) \defeq \frac{1}{n}\sum_{i=1}^n h(\langle w,
 x_i \rangle).
\end{align}

\noindent The following lemma, proved in
Appendix~\ref{sec:proof-lemma-neg-hinge}, shows that for a specific
$1$-Lipschitz continuous function $h$, approximately minimizing the
loss function is NP-hard.

\begin{lemma}
\label{lemma:hardness-neg-hinge}
Consider the optimization problem ~\eqref{eqn:hardnes-define-l} based
on the function $h(x) \defeq \min\{0,x\}$. It is NP-hard to
compute a vector $\what\in\R^d$ such that $\ltwos{\what}\leq 1$ and
$\ell(\what) \leq \ell(\wstar) + \frac{1}{(n+2)d}$.
\end{lemma}

We reduce the problem described by
Lemma~\ref{lemma:hardness-neg-hinge} to the problem of
Theorem~\ref{theorem:piecewise-linear-hard}. We do so by analyzing the
piecewise linear function
\begin{align}
\label{eqn:define-piecewise-linear}
h(x) \defeq \begin{cases} 1 & x \geq 1/2,\\ 0 & x \leq -1/2,\\ x + 1/2 &
  \mbox{otherwise}.
\end{cases}
\end{align}
With this choice of $h$, consider an instance of the problem
underlying Lemma~\ref{lemma:hardness-neg-hinge}, say
\begin{align}
\label{EqnGeneric}
\min_{\ltwos{w}\leq 1} g(w) \defeq \frac{1}{n}\sum_{i=1}^n \min\{0,
\langle w, x_i\rangle\},
\end{align}
and let $\wstar$ be a vector that achieves the above minimum.

Our next step is to construct a new problem in dimension $d + 1$, and
argue that any approximate solution to it also leads to an approximate
solution to the generic instance~\eqref{EqnGeneric}.  For each $i \in
[n]$, define the $(d+1)$-dimensional vector $x'_i
\defeq (x_i/\sqrt{2},1/\sqrt{2})$, as well as the quantities $u \defeq
- e_{d+1}/\sqrt{2}$ and $v \defeq e_{d+1}/(2\sqrt{2})$.  Here
$e_{d+1}$ represents the unit vector aligning with the $(d+1)$-th
coordinate.  For a $(d+1)$-dimensional weight vector $\wtil \in
\real^{d+1}$, consider the objective function
\begin{align*}
\elltil(\wtil) \defeq \frac{1}{13n} \Big( 6n h(\langle \wtil, u\rangle) +
6n h(\langle \wtil, v\rangle) + \sum_{i=1}^n h(\langle \wtil, x'_i\rangle)
\Big).
\end{align*}
If we decompose $\wtil$ as $\wtil = (\alpha/\sqrt{2},\tau/\sqrt{2})$
where $\alpha\in \R^d$ and $\tau\in \R$, then these parameters satisfy
$\ltwos{\alpha}^2 + \tau^2 \leq 2$.  We then have the equivalent
expression
\begin{align*}
  \elltil(\wtil) = \frac{1}{13n} \Big( 6n h(- \tau / 2) + 6n h(\tau/4)
  + \sum_{i=1}^n h(\langle \alpha, x_i\rangle/2 + \tau/2) \Big).
\end{align*}

\begin{lemma}
\label{LemGapprox}
Given any $\epsilon \in (0, \frac{1}{26})$, suppose that $\wtil =
\big(\frac{\alpha}{\sqrt{2}}, \frac{\tau}{\sqrt{2}} \big)$ is an
approximate minimizer of the function $\elltil$ with additive error
$\epsilon$, Then we have\footnote{In making this claim, we define
${\bf 0}/\ltwos{{\bf 0}} \defeq {\bf 0}$.}
\begin{align}
\label{eqn:g-approx-minimizer}
g\Big(\frac{\alpha}{\ltwos{\alpha}}\Big) \leq g(\wstar) + 26 \epsilon.
\end{align}
\end{lemma}

Thus, if there is a polynomial-time algorithm for minimizing function
$\elltil$, then there is a polynomial-time algorithm for minimizing
function $g$. Applying the hardness result for minimization of the
function $g$ (see Lemma~\ref{lemma:hardness-neg-hinge}) completes the
proof of the proposition. \\

\noindent It remains to prove the two lemmas, and we do so in the
following two subsections.


\subsection{Proof of Lemma~\ref{lemma:hardness-neg-hinge}}
\label{sec:proof-lemma-neg-hinge}

We reduce MAX-2-SAT to the minimization problem.  Given a MAX-2-SAT
instance, we construct a loss function $\ell$ so that if any algorithm
computes a vector $\what$ satisfying
\begin{align}
\label{eqn:hardness-what-constraint}
	\ell(\what) \leq \ell(\wstar) + \frac{1}{(2n+2)d},
\end{align}
then the vector $\what$ solves MAX-2-SAT.

First, we construct $n+1$ vectors in $\R^d$. Define the vector
$x_0 \defeq
\frac{1}{\sqrt{d}}{\bf 1}_d$, and for $i = 1, \ldots, \numobs$, 
the vectors $x_i \defeq \frac{1}{\sqrt{d}}x'_i$, where $x_i'\in \R^d$
is given by
\begin{align*}
x_{ij}' = \begin{cases} 1 & \mbox{if $z_i$ appears in $c_j$},\\ -1
  & \mbox{if $\neg z_i$ appears in $c_j$},\\ 0 & \mbox{otherwise}.
\end{cases}
\end{align*}
It is straightforward to verify that that $\ltwos{x_i}\leq 1$ for any
$i\in\{0,1,\dots,n\}$. We consider the following minimization problem:
\begin{align*}
\ell(w) = \frac{1}{2n+2} \sum_{i=0}^n \Big(\min\{0,\langle w, x_i
\rangle\} + \min\{0, \langle w, -x_i \rangle\}\Big).
\end{align*}
The goal is to find a vector $\wstar \in \real^d$ such that
$\ltwos{\wstar}\leq 1$ and it minimizes the function $\ell(w)$.

Notice that for every index $i$, at most one of $\min\{0,\langle w,
x_i \rangle\}$ and $\min\{ 0,\langle w, -x_i \rangle\}$ is
non-zero. Thus, we may write the minimization problem as
\begin{align}
\min_{\ltwos{w}\leq 1} (2n+2)\ell(w)  = \min_{\ltwos{w}\leq 1}
\sum_{i=0}^n \Big(\min_{\alpha_i\in\{-1,1\}} \langle w, \alpha_i x_i
\rangle\Big) & = \min_{\alpha\in\{-1,1\}^{n+1}} \min_{\ltwos{w}\leq 1}
\sum_{i=0}^n \langle w, \alpha_i x_i \rangle\nonumber\\ 
& =
\min_{\alpha\in\{-1,1\}^{n+1}} - \ltwo{\sum_{i=0}^n \alpha_i x_i} \nonumber \\
& = -
\left(\max_{\alpha\in\{-1,1\}^{n+1}} \sum_{j=1}^d \Big( \sum_{i=0}^n
\alpha_{i}x_{ij} \Big)^2\right)^{1/2}.
\label{eqn:transform-w-to-alpha}
\end{align}
We claim that maximizing $\sum_{j=1}^d ( \sum_{i=0}^n \alpha_{i}x_{ij}
)^2$ with respect to $\alpha$ is equivalent to maximizing the number
of satisfiable clauses.  In order to prove this claim, we consider an
arbitrary assignment to $\alpha$ to construct a solution to the
MAX-2-SAT problem. For $i=1,2,\dots,n$, let $z_i=\tt true$ if
$\alpha_i =
\alpha_0$, and let $z_i = \tt false$ if $\alpha_i = -\alpha_0$. With
this assignment, it is straightforward to verify the following: if the
clause $c_j$ is satisfied, then the value of $\sum_{i=0}^n
\alpha_{i}x_{ij}$ is either $3/\sqrt{d}$ or $-3/\sqrt{d}$. If the
clause is not satisfied, then the value of the expression is either
$1/\sqrt{d}$ or $-1/\sqrt{d}$. To summarize, we have
\begin{align}
\label{eqn:max-sat-eqn}
\sum_{j=1}^d \Big( \sum_{i=0}^n \alpha_{i}x_{ij} \Big)^2 = 1 + \frac{
  8 \times \mbox{(\# of satisfied clauses)}}{d}.
\end{align}
Thus, solving problem~\eqref{eqn:transform-w-to-alpha} determines the maximum number of satisfiable clauses:
\begin{align*}
\mbox{(max \# of satisfied clauses)} = \frac{d}{8}\Big(\big(
\min_{\ltwos{w}\leq 1} (2n+2)\ell(w) \big)^2- 1\Big).
\end{align*}
By examining equation~\eqref{eqn:transform-w-to-alpha}
and~\eqref{eqn:max-sat-eqn}, we find that the value of $(2n+2)\ell(w)$
ranges in $[-3,0]$. Thus, the MAX-2-SAT number is exactly determined
if $(2n+2)\ell(\what)$ is at most $1/d$ larger than the optimal
value. This optimality gap is guaranteed by
inequality~\eqref{eqn:hardness-what-constraint}, which completes the
reduction.


\subsection{Proof of Lemma~\ref{LemGapprox}}

In order to prove the claim, we first argue that $\tau\in [0,2]$. If
this inclusion does not hold, then it can be verified that $h(- \tau /
2) + h(\tau/4) \geq \frac{6}{13}$, which implies that
\begin{align*}
\elltil(\wtil) \geq \frac{6}{13} + \frac{1}{13n}\sum_{i=1}^n h(\langle
\alpha, x_i\rangle/2 + \tau/2) \geq \frac{12}{26}.
\end{align*}
However, the feasible assignment $(\alpha,\tau) = ({\bf 0},1)$
achieves function value $\frac{11}{26}$, which contradicts the
assumption that $\elltil(\wtil)$ is $\epsilon$-optimal.

Thus, we assume henceforth $\tau\in [0,2]$, and we can rewrite
$\elltil$ as
\begin{align*}
\elltil(\wtil) = \frac{3|\tau - 1| + 9}{26} +
\frac{1}{13n}\sum_{i=1}^n h(\langle \alpha, x_i\rangle/2 + \tau/2).
\end{align*}
Notice that $h(x+1/2)$ is lower bounded by $1+\min\{0,x\}$. Thus, we
have the lower bound
\begin{align}
\elltil(\wtil) & \geq \frac{3|\tau - 1|+9}{26} + \frac{1}{13} +
\frac{1}{13n}\sum_{i=1}^n \min\{0,\langle \alpha, x_i\rangle/2 +
(\tau-1)/2\} \nonumber \\ 
& \geq \frac{11}{26} + \frac{g(\alpha/2)}{13} + \frac{|\tau - 1|}{13},
	\label{eqn-lw-tau-zero-to-two}
\end{align}
where the last inequality uses the 1-Lipschitz continuous of
$\min\{0,x\}$. Notice that the function $g$ satisfies $g(\beta w) =
\beta g(w)$ for any scalar $\beta$. Thus, we can further lower bound
the right-hand side by
\begin{align}\label{eqn:g-alpha-2-decompose}
g(\alpha/2) = \frac{\ltwos{\alpha}}{2}\cdot
g\Big(\frac{\alpha}{\ltwos{\alpha}}\Big) \geq \frac{1}{2}
g\Big(\frac{\alpha}{\ltwos{\alpha}}\Big) -
\frac{\max\{0,\ltwos{\alpha}-1\}}{2},
\end{align}
where the last inequality uses the fact that
$g(\frac{\alpha}{\ltwos{\alpha}}) \geq - 1$. Note that inequality~\eqref{eqn:g-alpha-2-decompose} holds even for~$\alpha = {\bf 0}$ according the definition that ${\bf 0}/\ltwos{{\bf 0}} = {\bf 0}$.
If the quantity
$\ltwos{\alpha}-1$ is positive, then $\ltwos{\alpha} > 1$ and
consequently $\tau < 1$. Thus we have
\begin{align*}
\ltwos{\alpha}-1 < \ltwos{\alpha}^2 - 1 \leq 1-\tau^2 =
(1+\tau)(1-\tau) < 2(1 - \tau).
\end{align*}
Thus, we can lower bound $g(\alpha/2)$ by
$\frac{1}{2}g(\frac{\alpha}{\ltwos{\alpha}}) - |\tau -1|$. Combining
this bound with inequality~\eqref{eqn-lw-tau-zero-to-two}, we obtain
the lower bound
\begin{align}
\label{eqn:wp-lower-bound}
\ell(\wtil) \geq \frac{11}{26} +
\frac{1}{26}g\Big(\frac{\alpha}{\ltwos{\alpha}}\Big).
\end{align}
Note that the assignment $(\alpha,\tau) = (\wstar,1)$ is feasible. For
this assignment, it is straightforward to verify that the function
value is equal to $\frac{11}{26} + \frac{1}{26}g(\wstar)$. Using the
fact that $\wtil$ is an $\epsilon$-optimal solution, we have
$\ell(\wtil) \leq \frac{11}{26} + \frac{1}{26}g(\wstar) +
\epsilon$. This combining with inequality~\eqref{eqn:wp-lower-bound}
implies that $g\Big(\frac{\alpha}{\ltwos{\alpha}}\Big) \leq g(\wstar)
+ 26\epsilon$, which completes the proof.


\section{Proof of Theorem~\ref{theorem:nn-rand}}
\label{sec:proof-nn-rand}

Recalling the definition~\eqref{eqn:define-rademacher-complexity} of
the Rademacher complexity, the following lemma bounds the complexity
of $\nn_m$.

\begin{lemma}\label{lemma:rademacher}
The Rademacher complexity of $\nn_m$ is bounded as $R_k(\nn_m) \leq
\sqrt{\frac{q}{k}} \; B^m$.
\end{lemma}
\noindent See Appendix~\ref{sec:proof-rademacher} for the proof.\\

We study a single iteration of
Algorithm~\ref{alg:nn-rand}. Conditioning on any $\{(x'_j,y'_j)\}$,
define the quantity \mbox{ $G(f) \defeq \frac{1}{k}
\sum_{j=1}^k h(-y'_jf(x'_j))$.} Since the function $h$ is
\mbox{$L$-Lipschitz continuous,} we have
\begin{align}
\label{eqn:hg-minus-hf}
h(-y'_j g(x'_j)) - h(-y'_j \fstar(x'_j)) \leq L|g(x'_j) -
\fstar(x'_j)| \quad \mbox{for any $j\in[k]$}.
\end{align}
Given any function $f:\R^d\to \R$, we denote the vector
$(f(x'_1),\dots,f(x'_k))$ by $\varphi(f)$. For instance,
$\ltwos{\varphi(g)-\varphi(\fstar)}$ represents the $\ell_2$-norm
$(\sum_{j=1}^k |g(x'_j) -
\fstar(x'_j)|^2)^{1/2}$. Inequality~\eqref{eqn:hg-minus-hf} and the
Cauchy-Schwarz inequality implies
\begin{align*}
G(g) - G(\fstar) \leq \frac{L}{k}\sum_{j=1}^k |g(x'_j) - \fstar(x'_j)|
\leq \frac{L}{\sqrt{k}} \ltwos{\varphi(g)-\varphi(\fstar)}.
\end{align*}

\noindent We need one further auxiliary lemma:
\begin{lemma}
\label{LemGminus}
For any fixed function $\fstar \in \nn_m$, we have
\begin{align}
\label{claim:g-minus-fstar}
\ltwos{\varphi(g) - \varphi(\fstar)} \leq (2m-1) \epsilon \sqrt{k} B^m
\quad \mbox{with probability at least}\quad p_m \defeq
\left(\frac{\epsilon}{4}\right)^{k(s^m - 1)/(s-1)}.
\end{align}
\end{lemma}
\noindent See Appendix~\ref{AppLemGminus} for the proof of this
claim.\\

Let us now complete the proof of the theorem, using this two lemmas.
Lemma~\ref{lemma:rademacher-generalization} implies that
\begin{align*}
\E \Big[ \sup_{f\in \nn_m} \big| \ell(f) - G(f) \big|\Big] \leq 4 L
R_k(\nn_m).
\end{align*}
By Markov's inequality, we have $\sup_{f\in \nn_m} \big| \ell(f) -
G(f) \big| \leq 5LR_k(\nn_m)$ with probability at least $1/5$. This
event only depends on the choice of $\{(x'_j,y'_j)\}$. If it is true,
then we have
\begin{align*}
\ell(g) &\leq G(g) + 5LR_k(\nn_m) = G(\fstar) + (G(g) - G(\fstar)) +
5LR_k(\nn_m)\\ &\leq \ell(\fstar) +
\frac{L}{\sqrt{k}}\ltwos{\varphi(g) - \varphi(\fstar)} + 10LR_k(\nn_m).
\end{align*}
By Lemma~\ref{lemma:rademacher}, we have $R_k(\nn_m) \leq
\sqrt{\frac{q}{k}} \prod_{l=1}^m B_l$. Thus, setting $k =
q/\epsilon^2$, substituting the bound into
equation~\eqref{claim:g-minus-fstar} and simplifying yields
\begin{align*}
\ell(g) \leq \ell(\fstar) + (2m+9) \epsilon L B^m
\end{align*}
with probability at least $p_m/5$. If we repeat the procedure for $T =
(5/p_m)\log(1/\delta)$ times, then the desired bound holds with
probability at least $1-\delta$. The time complexity is obtained by
plugging in the choices of $(s,k,T)$.



\subsection{Proof of Lemma~\ref{lemma:rademacher}}
\label{sec:proof-rademacher}

We prove the claim by induction on the number of layers $m$. It is
known~\citep{kakade2009complexity} that $R_k(\nn_1) \leq
\sqrt{\frac{q}{k}} \: B$. Thus, the claim holds for the base case $m =
1$.  Now consider some $m > 1$, and assume that the claim holds for
$m-1$.  We then have
\begin{align*}
R_k(\nn_1) = \E \left[\sup_{f\in \nn_m} \frac{1}{k} \sum_{i=1}^k
  \varepsilon_i {f(x'_i)}\right],
\end{align*}
where $\varepsilon_1,\dots,\varepsilon_n$ are Rademacher variables. By
the definition of $\nn_m$, we may write the expression as
\begin{align*}
R_k(\nn_1) = \E\left[\sup_{f_1,\dots,f_d \in \nn_{m-1}} \frac{1}{k}
  \sum_{i=1}^n \varepsilon_i \sum_{j=1}^d w_j \sigma(f_j(x'_i))\right]
& = \E\left[\sup_{f_1,\dots,f_d\in \nn_{m-1}} \frac{1}{k} \sum_{j=1}^d
  w_j \sum_{i=1}^k \varepsilon_i \sigma(f_j(x'_i))\right] \\
& \leq B \E\left[\sup_{f\in \nn_{m-1}} \frac{1}{k} \sum_{i=1}^k
  \varepsilon_i \sigma(f(x'_i))\right] \\
& = B R_k(\sigma \circ \nn_{m-1}),
\end{align*}
where the inequality follows since $\lones{w} \leq B$. Since the
function $\sigma$ is 1-Lipschitz continuous, following the proof of
inequality~\eqref{eqn:radphi-to-radabs}, we have
\begin{align*}
R_k(\sigma \circ \nn_{m-1}) \leq R_k(\nn_{m-1}) \leq
\sqrt{\frac{q}{n}} \; B^m,
\end{align*}
which completes the proof.


\subsection{Proof of Lemma~\ref{LemGminus}}
\label{AppLemGminus}

We prove the claim by induction on the number of layers $m$.  If $m =
1$, then $\fstar$ is a linear function and $\varphi(\fstar) \in
[-B_1,B_1]^n$. Since $\varphi(g)$ minimizes the $\ell_2$-distance to
vector $u$, we have
\begin{align}
\label{eqn:g-minus-fstar}
\ltwos{\varphi(g) - \varphi(\fstar)} \leq \ltwos{\varphi(g) - u} +
\ltwos{\varphi(\fstar) - u} \leq 2\ltwos{\varphi(\fstar) - u}.
\end{align}
Since $u$ is drawn uniformly from $[-B,B]^k$, with probability at
least $(\frac{\epsilon}{4})^k$ we have \mbox{$\linftys{\varphi(\fstar)
- u}\leq \frac{\epsilon B}{2}$,} and consequently
\begin{align*}
\ltwos{\varphi(g) -\varphi(\fstar)} \leq \sqrt{k} 
\linftys{\varphi(g) - \varphi(\fstar)} \leq \epsilon \sqrt{k} B,
\end{align*}
which establishes the claim.

For $m > 1$, assume that the claim holds for $m-1$. Recall that
$\fstar/B$ is in the convex hull of $\sigma \circ \nn_{m-1}$ and every
function $f\in \sigma \circ \nn_{m-1}$ satisfies $\ltwos{\varphi(f)}
\leq \sqrt{k}$. By the Maurey-Barron-Jones lemma
(Lemma~\ref{lemma:mbj}), there exist $s$ functions in $\nn_{m-1}$, say
$\ftilde_1,\dots,\ftilde_s$, and a vector $w\in \R^s$ satisfying
$\lones{w} \leq B$ such that
\begin{align*}
\Big\| \sum_{j=1}^s w_j \sigma(\varphi(\ftilde_j)) - \varphi(\fstar)
\Big\|_2 \leq B \sqrt{\frac{k}{s}}.
\end{align*}
Let $\varphi(\ftilde) \defeq \sum_{j=1}^s
w_j\sigma(\varphi(\ftilde_j))$. If we chose $s = \left\lceil
\frac{1}{\epsilon^2} \right\rceil$, then we have
\begin{align}
\label{eqn:ftilde-minus-fstar}
\ltwos{\varphi(\ftilde) - \varphi(\fstar)} \leq \epsilon \sqrt{k} B.
\end{align}
Recall that the function $g$ satisfies $g = \sum_{j=1}^s v_j
\sigma\circ g_j$ for $g_1,\dots,g_s\in \nn_{m-1}$.  Using the
inductive hypothesis, we know that the following bound holds with
probability at least $p_{m-1}^s$:
\begin{align*}
\ltwos{\sigma(\varphi(g_j)) - \sigma(\varphi(\ftilde_j))} \leq
\ltwos{\varphi(g_j) - \varphi(\ftilde_j)} \leq (2m-3)\epsilon \sqrt{k}
B^{m-1} \quad \mbox{for any $j\in[s]$}.
\end{align*}
As a consequence, we have
\begin{align}
&\ltwo{\sum_{j=1}^s w_j\sigma(\varphi(g_j)) - \sum_{j=1}^s
    w_j\sigma(\varphi(\ftilde_j))} \leq \sum_{j=1}^s
    |w_j| \cdot \ltwos{\sigma(\varphi(g_j))
    - \sigma(\varphi(\ftilde_j))} \nonumber\\
    & \qquad \leq \lones{w} \cdot \max_{j\in[s]}\{\ltwos{\sigma(\varphi(g_j))
    - \sigma(\varphi(\ftilde_j))}\} \leq (2m-3)\sqrt{k}\epsilon
    B^{m}. \label{eqn:wg-minus-ftilde}
\end{align}
Finally, we bound the distance between $\sum_{j=1}^s
w_j\sigma(\varphi(g_j))$ and $\varphi(g)$. Following the proof of
inequality~\eqref{eqn:g-minus-fstar}, we obtain
\begin{align*}
\ltwo{\varphi(g) - \sum_{j=1}^s w_j\sigma(\varphi(g_j))} \leq 2
\ltwo{u - \sum_{j=1}^s w_j\sigma(\varphi(g_j))}.
\end{align*}
Note that $\sum_{j=1}^s w_j\sigma(\varphi(g_j)) \in [-B,B]^k$ and $u$
is uniformly drawn from $[-B,B]^k$. Thus, with probability at least $
(\frac{\epsilon}{4})^k$, we have
\begin{align}
\label{eqn:g-minus-wg}
\Big\| \varphi(g) - \sum_{j=1}^s w_j\sigma(\varphi(g_j)) \Big\|_2 \leq
\epsilon \sqrt{k} B.
\end{align}
Combining
inequalities~\eqref{eqn:ftilde-minus-fstar},~\eqref{eqn:wg-minus-ftilde}
and~\eqref{eqn:g-minus-wg} and using the fact that $B \geq 1$, we have
\begin{align*}
	\linfty{\varphi(g) - \varphi(\fstar)} \leq
        (2m-1) \epsilon \sqrt{k} B^m,
\end{align*}
with probability at least
\begin{align*}
 p_{m-1}^s \cdot \left(\frac{\epsilon}{4}\right)^k
 = \left(\frac{\epsilon}{4}\right)^{k\Big(\frac{s(s^{m-1}-1)}{s-1}+1\Big)}
 = \left(\frac{\epsilon}{4}\right)^{k(s^m-1)/(s-1)} = p_m,
\end{align*}
which completes the induction.


\section{Proof of Theorem~\ref{theorem:separable-data-error}}
\label{sec:proof-boosting}

\subsection{Proof of part (a)}

We first prove $\fhat\in \nn_m$. Notice that $\fhat = \sum_{t=1}^T
\frac{B}{2 b_T}\log(\frac{1 - \mu_t}{1 + \mu_t})\Deltahat_t$. Thus, if
\begin{align}\label{eqn:sum-gamma-T}
\sum_{t=1}^T \frac{B}{2 b_T} \left| \log(\frac{1 - \mu_t}{1 +
  \mu_t})\right| \leq B,
\end{align}
then we have $\fhat\in \nn_m$ by the definition of $\nn_m$. The
definition of $b_T$ makes sure that inequality~\eqref{eqn:sum-gamma-T}
holds. Thus, we have proved the claim. The time complexity is obtained
by plugging in the bound from Theorem~\ref{alg:nn-rand}.

It remains to establish the correctness of $\fhat$. We may write any
function $f\in\nn_m$ as
\begin{align*}
f(x) = \sum_{j=1}^d w_j \sigma(f_j(x)) \quad\mbox{where $w_j\geq 0$
  for all $j\in [d]$}.
\end{align*}
The constraints $w_j\geq 0$ are always satisfiable, otherwise since
$\sigma$ is an odd function we may write $w_j \sigma(f_j(x))$ as
$(-w_j) \sigma(-f_j(x))$ so that it satisfies the constraint. The
function $f_j$ or $-f_j$ belongs to the class $\nn_{m-1}$. We use the
following result by \cite{shalev2010equivalence}: Assume that there exists
$\fstar\in \nn_m$ which separate the data with margin~$\gamma$. Then
for any set of non-negative importance weights $\{\alpha_i\}_{i=1}^n$,
there is a function $f\in \nn_{m-1}$ such that $\sum_{i=1}^n \alpha_i
\sigma(-y_if(x_i)) \leq - \frac{\gamma}{B}$. This implies that, for
every $t\in[T]$, there is $f\in \nn_{m-1}$ such that
\begin{align*}
G_t(f) = \sum_{i=1}^n \alpha_{t,i} \sigma(-y_i f(x_i)) \leq
-\frac{\gamma}{B}.
\end{align*}
Hence, with probability at least $1-\delta$, the sequence
$\mu_1,\dots,\mu_T$ satisfies the relation
\begin{align}\label{eqn:gt-margin-bound}
\mu_t = G_t(\ghat_t) \leq -\frac{\gamma}{2B} \quad \mbox{for every
  $t\in [T]$}.
\end{align}
Algorithm~\ref{alg:nn-fw} is based on running AdaBoost for $T$
iterations. The analysis of AdaBoost by \cite{schapire1999improved}
guarantees that for any $\beta > 0$, we have
\begin{align*}
\frac{1}{n}\sum_{i=1}^n e^{- \beta}\indicator[-y_i f_T(x_i) \geq -
  \beta] \leq \frac{1}{n}\sum_{i=1}^n e^{-y_i f_T(x_i)} \leq
\exp\Big(-\frac{\sum_{t=1}^T \mu_t^2}{2}\Big).
\end{align*}
Thus, the fraction of data that cannot be separated by $f_T$ with
margin $\beta$ is bounded by $\exp(\beta
- \frac{\sum_{t=1}^T \mu_t^2}{8 B^2})$. If we choose
\begin{align*}
\beta \defeq \frac{\sum_{t=1}^T \mu_t^2}{2} - \log (n+1),
\end{align*}
then this fraction is bounded by $\frac{1}{n+1}$, meaning that all
points are separated by margin $\beta$. Recall that $\fhat$ is a
scaled version of $f_T$. As a consequence, all points are separated by
$\fhat$ with margin
\begin{align*}
\frac{B \beta}{b_T} = \frac{\sum_{t=1}^T \mu_t^2 - 2 \log
  (n+1)}{\frac{1}{B}\sum_{t=1}^T\log(\frac{1-\mu_t}{1+\mu_t})}.
\end{align*}
Since $\mu_t \geq - 1/2$, it is easy to verify that
$\log(\frac{1-\mu_t}{1+\mu_t}) \leq 4|\mu_t|$. Using this fact and
Jensen's inequality, we have
\begin{align*}
\frac{B \beta}{b_T} \geq \frac{(\sum_{t=1}^T |\mu_t|)^2/T - 2\log
  (n+1)}{ \frac{4}{B}\sum_{t=1}^T |\mu_t|}.
\end{align*}
The right-hand side is a monotonically increasing function of
$\sum_{t=1}^T |\mu_t|$. Plugging in the bound
in~\eqref{eqn:gt-margin-bound}, we find that
\begin{align*}
\frac{B \beta}{b_T} \geq \frac{\gamma^2 T/(4B^2) - 2\log(n+1)}{2
  \gamma T / B^2}.
\end{align*}
Plugging in $T = \frac{16B^2\log(n+1)}{\gamma^2}$, some algebra shows
that the right-hand side is equal to $\gamma/16$ which completes the
proof.


\subsection{Proof of part (b)}

Consider the empirical loss function $\ell(f) \defeq
\frac{1}{n}\sum_{i=1}^n h(-y_if(x_i))$, where $h(t) \defeq \max\{0, 1
+ 16 t / \gamma\}$.  Part (a) implies that $\ell(\fhat) = 0$ with
probability at least $1 - \delta$. Note that $h$ is
$(16/\gamma)$-Lipschitz continuous; the Rademacher complexity of
$\nn_m$ with respect to $n$ i.i.d.~samples is bounded by
$\sqrt{q/n}B^m$ (see Lemma~\ref{lemma:rademacher}). By the classical
Rademacher generalization bound~\cite[][Theorem 8 and Theorem
12]{bartlett2003rademacher}, if $(x,y)$ is randomly sampled form
$\mathbb{P}$, then we have
\[
	\E[h(-y\fhat(x))] \leq \ell(\fhat)
	+ \frac{32B^m}{\gamma} \cdot \sqrt{\frac{q}{n}}
	+ \sqrt{\frac{8\log(2/\delta)}{n}} \quad \mbox{with
	probabality at least $1-\delta$}.
\]
Thus, in order to bound the generalization loss by $\epsilon$ with
probability $1-2\delta$, it suffices to choose $n
= \poly(1/\epsilon,\log(1/\delta))$. Since $h(t)$ is an upper bound on
the zero-one loss $\indicator[t\geq 0]$, we obtain the claimed bound.


\section{Proof of Corollary~\ref{coro:flip-label}}
\label{sec:proof-flip-label}

The first step is to use the improper learning
algorithm~\cite[][Algorithm 1]{zhang2015ell_1} to learn a predictor
$\ghat$ that minimizes the following risk function:
\begin{align*}
\ell(g) \defeq \E[\phi(- \tildey g(x))] \quad \mbox{where} \quad \phi(t) \defeq 
	\left\{\begin{array}{ll} - \frac{2\eta}{1-2\eta} + \frac{\eta
	(t + \gamma)}{(1-\eta)(1-2\eta)\gamma} & \mbox{if $t \leq
	-\gamma$,}\\ - \frac{2\eta}{1-2\eta} + \frac{t
	+ \gamma}{(1-2\eta)\gamma} & \mbox{if $t >
	- \gamma$.}  \end{array}\right.  
\end{align*} 
Since $\eta < 1/2$, the function $\phi$ is convex and Lipschitz
continuous. The activation function ${\rm erf}(x)$ satisfies the
condition of~\cite[][Theorem 1]{zhang2015ell_1}. Thus, with sample
complexity $\poly(1/\tau,\log(1/\delta))$ and time complexity
$\poly(d,1/\tau,\log(1/\delta))$, the resulting predictor $\ghat$
satisfies
\begin{align*}
\ell(\ghat) \leq \ell(\fstar) + \tau \quad \mbox{with probability at least $1-\delta/3$}.
\end{align*}
By the definition of $\tildey$ and $\phi$, it is straightforward to
verify that
\begin{align}
\ell(g) = \E[(1-\eta)\phi(-yg(x)) + \eta \phi(yg(x))]
	= \E[\psi(-yg(x))] \label{eqn:lg-hinge}
        \end{align}
where
\begin{align*}
\psi(t) & \defeq 
\begin{cases}
0 & \mbox{if $t < -\gamma$,} \\ 
1 + t/\gamma & \mbox{if $-\gamma \le
	t \le \gamma$,} \\ 
2 + \frac{2\eta^2 - 2\eta + 1}{(1-\eta)(1-2\eta)\gamma}(t-\gamma)
	& \mbox{if $t > \gamma$.}  
\end{cases} 
\end{align*}
Recall that $y\fstar(x) \geq \gamma$ almost surely.  From the
definition of $\psi$, we have $\ell(\fstar) = 0$, so that
$\ell(\ghat) \leq \ell(\fstar) + \tau$ implies
$\ell(\ghat) \leq \tau$. Also note that $\psi(t)$ upper bounds the
indicator $\indicator[t\geq 0]$, so that the right-hand side of
equation~\eqref{eqn:lg-hinge} provides an upper bound on the
probability $\mprob(\sign(g(x))\neq y)$.  Consequently, defining
the classifier
$\hhat(x) \defeq \sign(g(x))$, then we have
\begin{align*}
\mprob( \hhat(x) \neq y) \leq \ell(\ghat) \leq \tau \quad \mbox{with 
probability at least $1-\delta/3$}.
\end{align*} 

Given the classifier $\hhat$, we draw another random dataset of $n$
points taking the form $\{(x_i, y_i)\}_{i=1}^n$. If $\tau
= \frac{\delta}{3n}$, then this dataset is equal to
$\{(x_i, \hhat(x_i))\}_{i=1}^n$ with probability at least
$1-2\delta/3$. Let the BoostNet algorithm take
$\{(x_i, \hhat(x_i))\}_{i=1}^n$ as its input. With sample size $n
= \poly(1/\epsilon,\log(1/\delta))$,
Theorem~\ref{theorem:separable-data-error} implies that the algorithm
learns a neural network $\fhat$ such that $\mprob({\rm
sign}(\fhat(x)) \neq y) \leq \epsilon$ with probability at
least~$1-\delta$. Plugging in the assignments of $n$ and $\tau$, the
overall sample complexity is $\poly(1/\epsilon,1/\delta)$ and the
overall computation complexity is $\poly(d,1/\epsilon,1/\delta)$.


\section{Proof of Proposition~\ref{theorem:hardness-neuralnet-margin}}
\label{sec:proof-hardness-neuralnet-margin}

We reduce the PAC learning of intersection of $T$ halfspaces to the
problem of learning a neural network. Assume that $T = \Theta(d^\rho)$
for some $\rho > 0$. We claim that for any number of pairs taking the
form $(x,h^*(x))$, there is a neural network $\fstar \in \nn_2$ that
separates all pairs with margin $\gamma$, and moreover that the margin
is bounded as $\gamma = 1/\poly(d)$.

To prove the claim, recall that $h^*(x) = 1$ if and only if $h_1(x)
= \dots = h_T(x) = 1$ for some $h_1,\dots,h_T\in H$. For any $h_t$,
the definition of $H$ implies that there is a $(w_t,b_t)$ pair such
that if $h_t(x) = 1$ then $w_t^T x - b_t - 1/2 \geq 1/2$, otherwise
$w_t^T x - b_t - 1/2 \leq -1/2$. We consider the two possible choices
of the activation function:

\begin{itemize}
\item {\bf Piecewise linear function:} 
If $\sigma(x) \defeq \min\{1,\max\{-1,x\}\}$, then let
\begin{align*}
g_t(x) \defeq \sigma( c(w_t^T x - b_t - 1/2) + 1),
\end{align*}
for some quantity $c > 0$. The term inside the activation function can
be written as $\langle \wtil, x'\rangle$ where
\begin{align*}
\wtil = (c\sqrt{2d+2}w_t, -c\sqrt{2d+2}(b_t+1/2), \sqrt{2}) \quad 
\mbox{and} \quad 
 x' =
 ( \frac{x}{\sqrt{2d+2}}, \frac{1}{\sqrt{2d+2}}, \frac{1}{\sqrt{2}}).
\end{align*}
Note that $\ltwos{x'}\leq 1$, and with a sufficiently small constant
$c = 1/\poly(d)$ we have $\ltwos{\wtil} \leq 2$. Thus, $g_t(x)$ is the
output of a one-layer neural network. If $h_t(x) = 1$, then $g_t(x) =
1$, otherwise $g_t(x) \leq 1 - c/2$. Now consider the two-layer neural
network $f(x) \defeq c/4 - T + \sum_{t=1}^T g_t(x)$. If $h^*(x) = 1$,
then we have $g_t(x) = 1$ for every $t\in [T]$ which implies $f(x) =
c/4$. If $h^*(x) = -1$, then we have $g_t(x) \leq 1- c/2$ for at least
one $t\in [T]$ which implies $f(x) \leq -c/4$. Thus, the neural
network $f$ separates the data with margin $c/4$. We normalize the
edge weights on the second layer to make $f$ belong to $\nn_2$. After
normalization, the network still has margin $1/\poly(d)$.

\item {\bf ReLU function:} if $\sigma(x) \defeq \max\{0,x\}$, then let
  $g_t(x) \defeq \sigma(-c(w_t^Tx - b_t-1/2))$ for some quantity $c >
  0$. We may write the term inside the activation function as $\langle
  \wtil, x'\rangle$ where $\wtil = (-c\sqrt{d+1} w_t, c\sqrt{d+1}
  (b_t+1/2))$ and $x' = (x,1)/\sqrt{d+1}$. It is straightforward to
  verify that $\ltwos{x'}\leq 1$, and with a sufficiently small $c =
  1/\poly(d)$ we have $\ltwos{\wtil} \leq 2$. Thus, $g_t(x)$ is the
  output of a one-layer neural network. If $h_t(x) = 1$, then $g_t(x)
  = 0$, otherwise $g_t(x) \geq c/2$. Let $f(x) \defeq c/4 -
  \sum_{t=1}^T g_t(x)$, then this two-layer neural network separates
  the data with margin $c/4$. After normalization the network belongs
  to $\nn_2$ and it still separates the data with margin $1/\poly(d)$.
\end{itemize}

To learn the intersection of $T$ halfspaces, we learn a neural network
based on $n$ i.i.d.~points taking the form $(x,h^*(x))$. Assume that
the neural network is efficiently learnable. Since there exists
$\fstar\in \nn_m$ which separates the data with margin $\gamma =
1/\poly(d)$, we can learn a network $\fhat$ in
$\poly(d,1/\epsilon,1/\delta)$ sample complexity and time complexity,
and satisfies $\mprob({\rm sign}(\fhat(x))\neq h^*(x))\leq \epsilon$
with probability $1-\delta$.  It contradicts with the assumption that
the intersection of $T$ halfspaces is not efficiently learnable.

\end{document}